\documentclass[acmtog]{acmart}

\usepackage{booktabs} % For formal tables

\usepackage[ruled]{algorithm2e} % For algorithms

\SetAlFnt{\small}
\SetAlCapFnt{\small}
\SetAlCapNameFnt{\small}
\SetAlCapHSkip{0pt}

\acmJournal{TOG}
\usepackage{booktabs}
\usepackage{multirow}
\usepackage{xcolor}
\usepackage{makecell}
\usepackage{rotating}
\usepackage{multirow}
\usepackage{graphicx}

\begin{document}
% Title portion
% \title{VecFontLLM: Anchor-Guided Multimodal Language Modeling for Direct Few-Shot Chinese Vector Font Synthesis}
\title{VecFontLLM: Anchor-Guided Direct Synthesis of Chinese Vector Fonts}

% DO NOT ENTER AUTHOR INFORMATION FOR ANONYMOUS TECHNICAL PAPER SUBMISSIONS TO SIGGRAPH 2019!
\author{Hao Yuan}
% \orcid{1234-5678-9012-3456}
\affiliation{%
 \institution{Fuzhou University}
%  \streetaddress{104 Jamestown Rd}
%  \city{Williamsburg}
%  \state{VA}
%  \postcode{23185}
 \country{China}
}
% \email{gang_zhou@wm.edu}

\author{Yuxuan Luo}
\affiliation{%
 \institution{Peking University}
 % \city{Rocquencourt}
 \country{China}
}
% \email{beranger@inria.fr}

\author{Xing Chen}
\affiliation{%
 \institution{Fuzhou University}
% \streetaddress{Rono-Hills}
% \city{Doimukh}
% \state{Arunachal Pradesh}
 \country{China}
}
% \email{aprna_patel@rguhs.ac.in}
\author{Zhouhui Lian}
\affiliation{%
 \institution{Peking University}
 % \streetaddress{30 Shuangqing Rd}
 % \city{Haidian Qu}
 % \state{Beijing Shi}
 \country{China}
}
% \email{chan0345@tsinghua.edu.cn}

%\renewcommand\shortauthors{Zhou, G. et al}

\begin{abstract}
% Vector glyphs are central to digital font files and costly to design, yet automatic Chinese vector font synthesis from limited style references remains unresolved.
% Existing vector font generation methods lack robust modeling of vector representations, while recent LLM-based approaches focus primarily on alignment tasks in vector graphics.
% Furthermore, existing methods overlook both the inherent structural complexity of vector representations and the challenges posed by the long and intricate drawing instruction sequences of Chinese glyphs.
% To address these limitations, we propose VecFontLLM, a multimodal large language model (MLLM) for high-quality Chinese vector font synthesis. 
% The key technical contributions are threefold: 
% An MLLM with a dedicated vector font style encoder that directly generates vector glyphs from a small set of reference vector glyphs;
% A vector font synthesis pipeline with command–coordinate decoupling to alleviate the modeling complexity of vector glyph generation;
% A confidence-guided test-time scaling strategy adapted to the compositional structure of Chinese glyphs, improving stability in long sequence generation.
% Using our model, \textbf{for the first time, high-quality vector fonts consisting of arbitrarily complex Chinese glyphs can be directly synthesized} without relying on image synthesis models. 
% Both qualitative and quantitative experiments have been conducted to demonstrate the superiority of VecFontLLM compared to existing vector font synthesis methods.

Direct generation of Chinese vector fonts is a challenging and ongoing problem. 
A Chinese vector glyph contains complex component structure, anchor layout, and Bézier curve details, which work at different scales, but a standard vector sequence writes them together in one long sequence, making the task of vector font synthesis challenging.
Existing direct vector generators often fail on complex characters, while raster-domain methods must vectorize the synthesized glyph images afterward.
To address the above-mentioned problem, this paper proposes VecFontLLM, an anchor-guided multimodal large language model for direct few-shot synthesis of Chinese vector fonts. 
Our key idea is to generate vector glyphs through anchors rather than a standard vector sequence. 
Specifically, the proposed VecFontLLM first predicts and refines an anchor scaffold that fixes the coarse layout of components and contours, and then completes Bézier control points to recover local curvature and style. 
At test time, a confidence-guided generation chain samples multiple component candidates and continues synthesis from the highest-confidence one, improving stability for complex glyphs.
This work demonstrates, \textbf{for the first time, high-quality few-shot synthesis of complex Chinese vector glyphs directly in the vector domain}, without raster generation or vectorization. 
Experiments on several Chinese font datasets show substantial improvements over existing vector font synthesis methods, competitive glyph rendering quality against raster-domain baselines, and vector command distributions close to real fonts.
\end{abstract}


%
% End generated code
%


\begin{teaserfigure}
  \includegraphics[width=\textwidth]{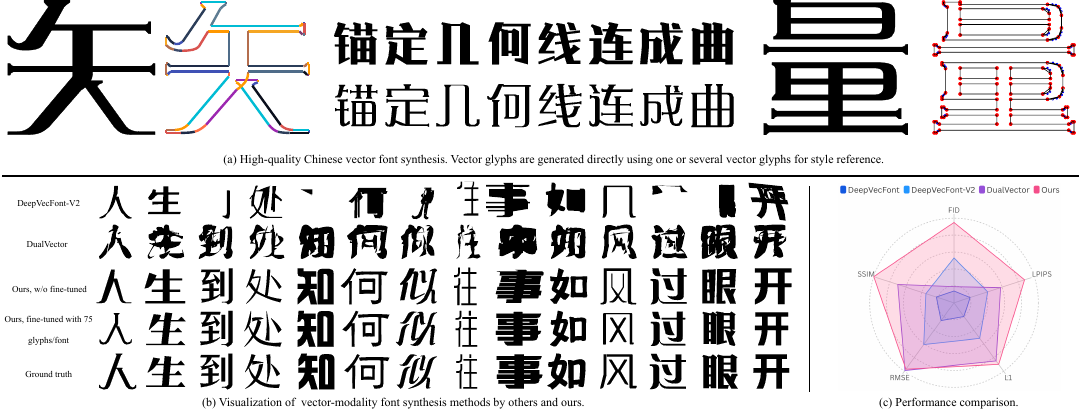}
  \caption{VecFontLLM is an MLLM-based framework optimized for high-fidelity Chinese vector font synthesis (a). Compared with existing vector font synthesis methods, which fail to produce recognizable results (b). VecFontLLM achieves state-of-the-art performance across multiple evaluation metrics (c).}
    \Description{
    Overview of VecFontLLM framework for Chinese vector font synthesis.
    }
  \label{fig0}
\end{teaserfigure}

\maketitle

\section{Introduction}
\label{sec:intro}

Vector fonts are the standard representation of digital typefaces in design and production workflows.
Unlike raster images, they describe each glyph as editable outlines, where contours are specified by path commands and Bézier control points.
This representation is compact and resolution independent, but it is difficult for font synthesis: the output must be valid and editable vector graphs (typically in the SVG format), rather than merely structured raster images.

Most of the recent font generation methods work in the raster domain and produce visually pleasing results~\cite{mxfont,cggan,fontdiffuser,hfhfont,beyondpatches}.
However, this formulation largely avoids the constraints of vector fonts. 
Vector outlines are harder to synthesize because topology, path commands, anchors, and curve geometry are tightly coupled.
Commands define path segments and their order, anchors set the layout, and Bézier control points shape local curvature. 
When all of them are predicted in one vector sequence, these different geometric roles become tightly coupled.
If an early command changes the contour structure, later coordinates and control points follow the wrong topology. 
Local Bézier control errors may keep the topology unchanged, but they can still break alignment or curve continuity.
Thus, vector font synthesis is treated not as a mere raster matching task, but as the generation of rigorous geometric representations.
% structurally consistent geometric representation.

% Many existing vector-domain font generators~\cite{deepvecfont,deepvecfontv2,vecfusion,diffvecfont,vectorglyph} and MLLM-based SVG synthesis methods~\cite{empowering,iconshop,omnisvg,starvector} adopt a raw SVG sequence.
Many existing vector-domain font generators~\cite{deepvecfont,deepvecfontv2,vecfusion,diffvecfont,vectorglyph} represent glyphs as raw SVG sequences.
Recent multimodal large language model (MLLM)-based SVG synthesis methods~\cite{empowering,iconshop,omnisvg,starvector} use a similar formulation.
% However, a glyph outline is not built from a single token sequence. 
% Contour organization, anchor placement, and curve shaping play different geometric roles.
A single sequence must encode contour organization, anchor placement, and curve shaping, although they play different geometric roles.
The first governs whether the validity of the glyph structure, the second determines the coarse layout of closed paths, and the third controls local curvature and stylistic details. 
These factors are interleaved in the final path representation, but treating them as one sequence hides their different roles and makes errors hard to localize.

This issue is amplified in Chinese vector font synthesis. 
Chinese glyphs are highly compositional.
A character often contains multiple visual components arranged under specific spatial relations. 
In vector form, each such component may correspond to a group of closed contours, including holes or nested interior paths. 
% They are also curve-rich, as terminals, bending trends, and stroke modulation must be captured through Bézier geometry rather than pixel-level appearance. 
They also contain many curved stroke details, such as stroke terminals, bends, and thickness changes, which must be represented by Bézier control points.
In our few-shot setting, the model must preserve the component layout of unseen characters while transferring such local curve patterns from only a few reference glyphs. 
A single SVG sequence predicts component structure, anchor placement, and curve parameters, so early errors can limit later geometric results.

To avoid coupling these predictions in one SVG sequence, we introduce VecFontLLM, a vector-conditioned MLLM architecture for few-shot Chinese vector font synthesis.
VecFontLLM combines an image-content encoder with a vector-style encoder, so that SVG generation is conditioned on both the target character and a small set of vector style exemplars. 
Rather than predicting the final SVG path, the model first constructs and refines an anchor scaffold, where anchors correspond to the endpoints of path segments and capture the coarse layout of components and contours. 
Conditioned on the refined scaffold, it then completes Bézier control points to synthesize local curvature and style details. 
For complex Chinese characters, we further use confidence-guided test-time generation to make each generated component more reliable.
The model generates several component candidates, scores them by confidence, and selects the best one before moving to the next component.

We evaluate VecFontLLM on several Chinese font datasets against both vector-domain generators and raster-based font synthesis methods. 
Fig.~\ref{fig0}(b) shows that our method better preserves component structure and local stroke style than previous direct vector baselines. 
Fig.~\ref{fig0}(c) shows the same trend in perceptual and reconstruction metrics.
Against raster-domain font generators, VecFontLLM achieves competitive rendered quality while retaining vector outputs that are resolution-independent and directly editable. 
Our ablations further show that the anchor-to-curve generation improves the baseline, and that the confidence-guided generation chain further stabilizes complex glyph generation.
The command-distribution analysis also suggests that the command statistics of generated glyphs remain close to those of real vector fonts, rather than merely matching their rasterized appearance.

Major contributions of this paper are summarized as follows:
\begin{itemize}

    \item We design VecFontLLM, a vector-based MLLM architecture that integrates target-content encoding and vector-style encoding with a language model for vector glyph generation.

    \item We introduce an anchor-guided generation process that first constructs and refines an anchor scaffold, and then completes Bézier control points to synthesize curve geometry.

    \item We propose a confidence-guided test-time scaling strategy for complex Chinese glyphs, where a generation chain samples multiple component candidates at each step and continues synthesis from the highest-confidence one. 

    \item Extensive experiments demonstrate the superiority of our method to the state of the art. With the proposed VecFontLLM, for the first time, high-quality complex Chinese vector glyphs can be directly synthesized. 
    
    %We advance few-shot Chinese font synthesis in the native vector domain, producing recognizable and editable SVG glyphs for complex characters from a small number of vector style exemplars, without relying on vectorization.
    
\end{itemize}

\section{Related Work}
\label{sec:Related Work}

\paragraph{Font generation.}
Font synthesis spans two major modalities: image and vector.
In the image modality, earlier studies ~\cite{zigan,cggan,dmfont,strokegan,dgfont} adopted GAN- and VAE-based generative frameworks ~\cite{gan,vae}, while more recent work ~\cite{fontdiffuser,difffont,hfhfont} exploits the stronger generative capabilities of diffusion models ~\cite{diffusion,ltm}.
However, the image modality is inherently unsuitable for practical font design. 
Although HFH-Font ~\cite{hfhfont} produces vector fonts via a super-resolution model followed by vectorization, the process is time-consuming and lacks precision.
In the vector modality, early research ~\cite{sketchrnn,deepsvg,svgvae} focused on learning SVG encodings and corresponding generative frameworks. 
Subsequent works such as DeepVecFont ~\cite{deepvecfont} and DeepVecFont-v2 ~\cite{deepvecfontv2} leverage VAE-based and dual-modality architectures, while VecFusion ~\cite{vecfusion} and DiffVecFont ~\cite{diffvecfont} employ diffusion models for improved performance.
However, these methods struggle to generate Chinese vector fonts with high visual quality due to the inherent difficulty of vector representations and the limited capability of existing models.

\paragraph{General vector generation methods.}
Beyond the aforementioned vector font synthesis approaches, several studies have investigated more general vector graphic generation tasks.
To synthesize vector images that closely resemble raster ones, some works ~\cite{vecfusion,svgdreamer,text2vector,nivel} adopt Score Distillation Sampling (SDS)–inspired optimization techniques ~\cite{sds} for generating complex vector graphics.
However, these methods are computationally intensive and thus not ideal for efficient vector generation.
Therefore, recent research has increasingly shifted toward vector generation using LLMs and MLLMs.
Early works ~\cite{svgeditbench,vgbench,chat2svg} quantified the basic SVG-generation capability of LLMs, followed by subsequent studies ~\cite{iconshop,starvector,empowering,omnisvg} that strengthened SVG synthesis through SFT.
More advanced works explored Group Relative Policy Optimization (GRPO) ~\cite{deepseek} to enhance MLLM-based SVG generation ~\cite{rlrf4svg,reasonsvg,svgen} and investigated Chain-of-Thought reasoning for improved vector generation ~\cite{reasonsvg,svgen,svgthinker}.
In addition, other efforts extended the task to SVG completion and understanding ~\cite{robosvg,internsvg}.
Despite this progress, these works remain largely focused on aligning text or bitmap modalities with vector data and still provide limited support for style transfer.

\paragraph{Training and inference of MLLMs.} 
MLLM-based vector generation ~\cite{starvector,omnisvg,reasonsvg} and font or character recognition methods ~\cite{callireader,deepseekocr} mainly enhance model capabilities via training techniques such as SFT and GRPO, while largely neglecting the reasoning potential of LLMs.
In mathematics reasoning, methods like Self-Consistency ~\cite{self-consistency}, Adaptive-Consistency ~\cite{adaptive-consistency}, and Soft-Self-Consistency ~\cite{soft-self-consistencey} select the most reliable answer from multiple sampled reasoning paths.
Due to the high cost of sequence-level parallel reasoning, recent research focuses on fine-grained evaluation. Techniques such as Claim-Conditioned Probability ~\cite{fact-checking}, Self-Certainty ~\cite{self-certainty}, and Deep Think with Confidence ~\cite{deepconf} measure token-level reliability to improve accuracy and stability.
However, such test-time scaling strategies remain unexplored for LLM-based vector generation.

\section{Method Description}
\label{sec:method}

\begin{figure*}
  \centering
    \includegraphics[width=1.0\linewidth]{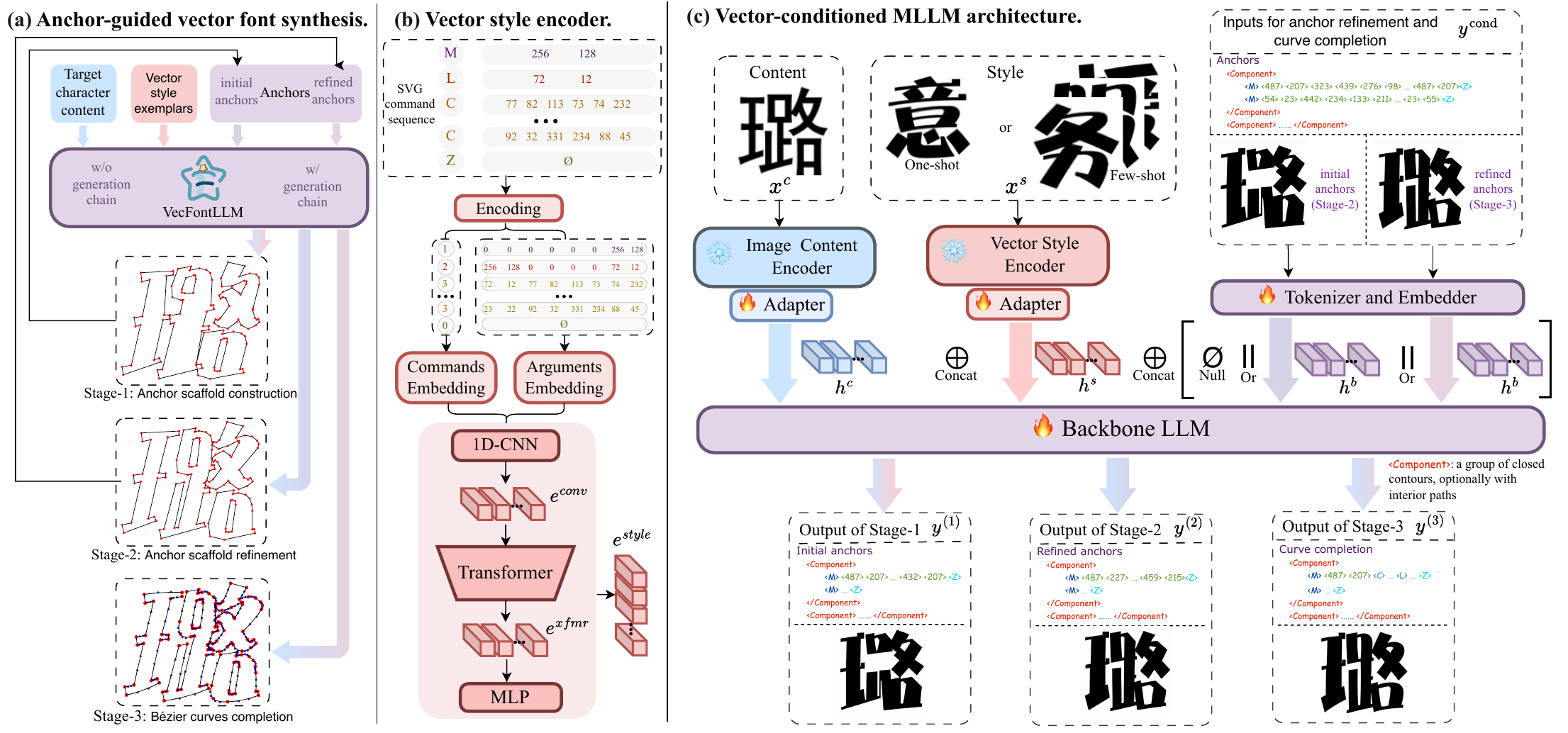}
    \caption{Overview of VecFontLLM. (a) We factorize Chinese vector font synthesis into anchor scaffold construction, anchor scaffold refinement, and Bézier curve completion. A confidence-guided generation chain selects high-confidence component candidates during inference for complex glyphs. (b) The vector style encoder converts SVG command and argument sequences into a vector-style representation using command/argument embeddings, a 1D-CNN, and a Transformer. (c) In the vector-conditioned MLLM, target content and vector style exemplars are encoded separately and fed to a backbone LLM, which produces stage-specific vector tokens for initial anchors, refined anchors, and completed curves.}
    \label{fig2}
\end{figure*}

In this section, we describe VecFontLLM, an anchor-guided framework for direct Chinese vector font synthesis. 
Section~\ref{method:pipeline} describes the three stages of synthesis, including scaffold construction, anchor refinement, and Bézier curve completion.
Section~\ref{method:vecfontllm} then details the vector-conditioned MLLM architecture, including the vector style encoder, the content and style conditioning scheme, and the training objectives. 
Finally, Section~\ref{sec:test_time} describes the confidence-guided generation chain used at test time to select reliable component candidates for complex glyphs.

\subsection{The Pipeline of VecFontLLM}
\label{method:pipeline}

VecFontLLM generates a vector glyph through an anchor-guided pipeline instead of predicting the complete SVG path stream in one pass.
As shown in Fig.~\ref{fig2}(a), the pipeline consists of three stages: anchor scaffold construction, anchor scaffold refinement, and Bézier curve completion.
We refer to the endpoint, i.e., the terminal coordinate, of each path segment as an anchor.
Anchors define the coarse layout of components and contours, while Bézier control points determine local curvature and style.

In Stage-1, the model predicts an initial anchor scaffold by generating the anchors of path segments, omitting Bézier control points.
Connecting these anchors gives a coarse polygonal outline that captures the main structure of the glyph.
In Stage-2, the initial scaffold is fed back to the model to refine anchor placement and correct errors in the coarse layout.
In Stage-3, the model completes Bézier control points conditioned on the refined scaffold, producing the final SVG path with local curvature.
All stages share the same VecFontLLM architecture and are fine-tuned with stage-specific data.

This decomposition follows the different roles of endpoints and control points in vector outlines.
In a single SVG path, anchor placement and curve parameters are interleaved in one token sequence.
For cubic Bézier segments, endpoints define the global outline layout.
Control points determine local curvature and style.
By constructing and refining an anchor scaffold before curve completion, VecFontLLM separates coarse outline layout from local Bézier geometry while remaining in the native vector representation.
During inference, the confidence-guided generation chain can be used during scaffold generation for complex glyphs.

\subsection{The Architecture and Training of VecFontLLM}
\label{method:vecfontllm}

We next describe VecFontLLM, the vector-conditioned MLLM used across the three stages of our synthesis pipeline.
Given a reference glyph image and a few vector style exemplars, VecFontLLM encodes content and style separately, then projects both features into the LLM embedding space.
All three stages use the same architecture, but each is fine-tuned with its own input and output format.

\subsubsection{Vector Style Encoder}
\label{vfs}

\paragraph{Vector glyph representation.}
Following DeepVecFont-V2~\cite{deepvecfontv2}, we represent a vector glyph as a sequence of drawing instructions \(X=[I_1,I_2,\ldots,I_N]\), where each instruction \(I_i=(C_i,A_i)\) consists of a command type \(C_i\in\{\texttt{M},\texttt{L},\texttt{C},\texttt{Z}\}\) and its coordinate arguments \(A_i\). 
We treat each SVG instruction as a path segment and include its starting point in the arguments. 
A cubic Bézier segment therefore has four coordinate pairs: the starting point, two control points, and the endpoint.
Commands with fewer valid coordinates are zero-padded to the fixed-length form \(A_i=[(x_i^1,y_i^1),\ldots,(x_i^4,y_i^4)]\). 
More details are provided in the Appendix.

We embed commands and coordinate arguments separately. 
Each command \(C_i\) and coordinate value \(x_i^j,y_i^j\) is mapped to a \(d_E\)-dimensional embedding, denoted by \(e_i^C\), \(e_i^{x^j}\), and \(e_i^{y^j}\). 
The coordinate embeddings are concatenated and projected to form the argument embedding:
\begin{equation}
  e_i^{A} = W^A
  \left(
  e_i^{x^1} \oplus e_i^{y^1} \oplus \cdots \oplus e_i^{x^4} \oplus e_i^{y^4}
  \right),
  \label{eq:arg_embedding}
\end{equation}
where \(\oplus\) denotes concatenation and \(W^A \in \mathbb{R}^{d_E \times 8d_E}\). 
The final instruction embedding is
\begin{equation}
  e_i^I = e_i^C + e_i^A + e_i^{pos},
  \label{eq:inst_embedding}
\end{equation}
where \(e_i^{pos}\) is the positional embedding.

\paragraph{Style encoding and contrastive training.}
As shown in Fig.~\ref{fig2}(b), the instruction embeddings are fed into the vector style encoder to obtain a font-level vector-style feature.
Each instruction is first embedded from its command and coordinate arguments. 
The encoder then uses a 1D-CNN~\cite{cnn} to aggregate local instruction patterns and reduce the sequence length.
A Transformer~\cite{transformer} then models dependencies across path segments and contours.
Finally, an MLP projects the encoded sequence into the style feature \(e^{\mathrm{style}}\) for given vector glyphs.

We train the vector style encoder with a supervised contrastive objective~\cite{supcon}.
Positive pairs are formed by sampling two different glyphs from the same font, while glyphs from different fonts serve as negatives.
% This encourages the encoder to capture style cues shared across characters, rather than memorizing the content structure of a particular glyph.
This guides the encoder toward style cues shared across characters, instead of glyph-specific structure.

\subsubsection{Vector-Conditioned MLLM}
\label{vfllm}

\paragraph{Content and style conditioning.}
VecFontLLM uses StarCoder-1B~\cite{starcoder} as the backbone LLM and conditions it on a reference character image \(x^c\) and vector style exemplars \(x^s\).
% A frozen image encoder and the vector style encoder extract content and style features, respectively, which are projected by lightweight adapters into the language- model embedding space as conditioning embeddings \(h^c\) and \(h^s\).
A frozen image encoder extracts content features, and the vector encoder extracts style features. 
Lightweight adapters project both into the LLM embedding space as \(h^c\) and \(h^s\).
They are combined with the stage input embeddings to form the prefix context for autoregressive generation.

\paragraph{Stage-specific autoregressive training.}
VecFontLLM is trained autoregressively to predict stage-specific vector sequences.
Before stage fine-tuning, we warm up the backbone with a curriculum.
The curriculum moves from standard SVG code generation to our SVG token format, with larger data scale and higher coordinate precision in later stages. 
Details are given in the Appendix.
After this initialization, we fine-tune three separate models for anchor scaffold construction, anchor refinement, and Bézier curve completion.

We now formalize the autoregressive objectives for the three stages.
Let \(y^{(1)}\), \(y^{(2)}\), and \(y^{(3)}\) denote the target sequences for Stage-1, Stage-2, and Stage-3, respectively.
Here, \(y^{(1)}\) denotes the initial anchor scaffold, \(y^{(2)}\) the refined scaffold, and \(y^{(3)}\) the final vector sequence with completed Bézier curves.
For Stage-1, the model has no scaffold input. 
It predicts the initial anchor scaffold from the target content and style exemplars:
\begin{equation}
  p\left(y^{(1)} \mid x^c, x^s \right)
  =
  \prod_{i=1}^{L_1}
  p\left(
  y^{(1)}_i
  \mid
  y^{(1)}_{<i}, x^c, x^s
  \right).
  \label{eq:stage1_prob}
\end{equation}

For Stage-2 and Stage-3, the model additionally conditions on a conditioning sequence \(y^{\mathrm{cond}}\), which provides the scaffold to be refined or completed.
In Stage-2, \(y^{\mathrm{cond}}=y^{(1)}\), and the target is the refined scaffold \(y^{(2)}\).
In Stage-3, \(y^{\mathrm{cond}}=y^{(2)}\), and the target is the completed Bézier sequence \(y^{(3)}\).
The conditional distribution is
\begin{equation}
  p\left(y^{(s)} \mid x^c, x^s, y^{\mathrm{cond}} \right)
  =
  \prod_{i=1}^{L_s}
  p\left(
  y^{(s)}_i
  \mid
  y^{(s)}_{<i}, x^c, x^s, y^{\mathrm{cond}}
  \right),
  \qquad s \in \{2,3\}.
  \label{eq:stage23_prob}
\end{equation}

All stage-specific models are optimized with the standard autoregressive cross-entropy loss over their corresponding sequences.

\subsection{Confidence-guided Test-time Scaling}
\label{sec:test_time}

Complex Chinese glyphs often contain many components with closed or nested contours. 
Stage-1 therefore generates the scaffold component by component.
An early component error can affect subsequent components, and an unreliable anchor scaffold can further degrade anchor refinement and Bézier curve completion.
We use test-time scaling as confidence-guided selection in Stage-1, where structural errors have the largest downstream effect. 
The selection is performed at either the glyph level or the component level.

\paragraph{Confidence scoring.}
We assign a confidence score to each candidate sequence \(z=[z_1,\ldots,z_L]\), where \(z\) can be either a complete glyph or a single component.
For each generated token \(z_i\), the LLM outputs a probability distribution over the vocabulary, from which we compute the token confidence \(C_i\).
We evaluate four token-level measures: Gini impurity~\cite{gi}, entropy~\cite{entropy}, distributional perplexity~\cite{self-certainty}, and token confidence~\cite{tc}, with definitions given in the Appendix.
% All measures are rescaled so that larger values indicate higher confidence. 

Let \(C_{(i)}\) and \(C_{[i]}\) denote the \(i\)-th lowest and highest confidence scores after sorting \(\{C_i\}_{i=1}^{L}\), respectively.
The candidate confidence \(C(z)\) is obtained by aggregating token confidences:
\begin{align}
C^{\mathrm{mean}}(z) &= \frac{1}{L}\sum_{i=1}^{L} C_i, \label{eq:mean} \\
C^{\mathrm{bottom}}(z) &= \frac{1}{B}\sum_{i=1}^{B} C_{(i)}, 
\qquad B=\left\lceil b\% \cdot L \right\rceil, \label{eq:bottom} \\
C^{\mathrm{top}}(z) &= \frac{1}{T}\sum_{i=1}^{T} C_{[i]}, 
\qquad T=\left\lceil t\% \cdot L \right\rceil. \label{eq:top}
\end{align}

\paragraph{Glyph-level selection.}
A straightforward use of the candidate confidence \(C(z)\) is to treat the entire glyph as a candidate sequence: given \(K\) sampled glyphs \(\{z_k\}_{k=1}^{K}\), we select \(z^{*}=\arg\max_{z_k} C(z_k)\).
An illustration is provided in the Appendix.
We refer to this complete-glyph selection baseline as \emph{parallel thinking}; 
This strategy tests whether the confidence score can identify better vector generations and, when used in Stage-1, can select a more reliable scaffold for later anchor refinement and Bézier curve completion.
However, the score is computed only after the full glyph is generated. 
In complex Chinese glyphs, an early component error can corrupt the remaining scaffold, motivating the component-level strategy below.

\paragraph{Component-level generation chain.}
We address this delay by applying confidence-guided selection at the component level.
This choice follows the topology of Chinese vector glyphs: a \texttt{<component>} denotes one outer closed contour together with its associated interior contours, if any; see Appendix for an illustration.
Stage-1 scaffold construction proceeds component by component. 
At each step, the model samples candidate components from the current partial scaffold, scores them with \(C(z)\), and appends the highest-confidence one.
This process repeats until the end-of-sequence token is produced, allowing the generation chain to reject uncertain local structures before they constrain the remaining scaffold.

\section{Experiments}
\label{sec:experiment}

\subsection{Experimental Setup}
\label{experimentsetup}
\subsubsection{Datasets}
\label{dataset}

We evaluate Chinese font synthesis in both vector and rasterized image domains. 
Our font collection is divided into three groups: 
\textbf{Basic Fonts (BFs)}, containing 345 fonts with 6,763 Chinese glyphs each; 
\textbf{Unseen Fonts (UFFs)}, containing 15 fonts with the same 6,763 glyphs; 
and \textbf{Unseen Character Fonts (UCFs)}, containing 4 font sets with 20,964 Chinese glyphs each. 
All fonts are filtered to satisfy our vector-quality and sequence-length criteria.

We construct the following splits:
\begin{itemize}
  \item Train Set: 80\% of BFs for training and 10\% for validation.
  \item Base Test Set (BTS): the remaining 10\% of \textbf{BFs}, used for standard evaluation.
  \item Unseen Font Test Set (UFTS): \textbf{UFFs}, used to evaluate generalization to unseen font styles.
  \item Unseen Character Test Set (UCTS): \textbf{UCFs}, used to evaluate generalization to unseen character structures.
  \item Hard Base Test Set (HBTS): 6,582 failure cases produced by the Stage-1 model on BTS for test-time scaling evaluation.
  \item Small Test Set (STS): 495 simple characters from \textbf{UFFs}, used for comparison with prior vector-domain methods that cannot reliably handle complex Chinese glyph sequences.
\end{itemize}

\subsubsection{Evaluation Metrics and Implementation Details}

To compare vector- and raster-domain methods under the same rendered-glyph metrics, we rasterize all generated SVGs using CairoSVG~\cite{cairo}. 
We report FID~\cite{fid}, LPIPS~\cite{lpips}, L1, RMSE, and SSIM~\cite{ssim}. 
FID is computed using 2048-dimensional Inception-V3 features~\cite{inceptionv3}, and LPIPS is computed with AlexNet features~\cite{alexnet}.

We evaluate glyphs on the full canvas without cropping or alignment, since size, centering, and margins are part of the font style.
This preserves layout errors during metric computation.
Our method can be rasterized at \(512\times512\), whereas several baselines output lower-resolution glyphs.
For fair image-based evaluation, we upsample low-resolution predictions and apply the same downsample--upsample procedure to the ground truth.

\begin{table}[t]
  \centering
  \captionof{table}{Quantitative evaluation of vector-modality methods.}
  \label{vsVec}
  \resizebox{\linewidth}{!}{
    \begin{tabular}{@{}l|ccccc@{}}
      \toprule
      \multirow{2}{*}{Method} & \multicolumn{4}{c}{STS} \\
      & FID $\downarrow$ & LPIPS $\downarrow$ & L1 $\downarrow$ & RMSE $\downarrow$ & SSIM $\uparrow$ \\
      \midrule
      DeepVecFont & 217.784 & 0.587 & 0.344 & 0.550 & 0.512 \\
      DeepVecFont-V2 & 94.691 & 0.517 & 0.286 & 0.490 & 0.546 \\
      DualVector & 203.170 & 0.464 & 0.232 & \textbf{0.432} & 0.624 \\
      Ours($n_{\mathrm{ref}}=1$) & \underline{14.973} & \underline{0.379} & \underline{0.213} & 0.442 & \underline{0.679} \\
      Ours($n_{\mathrm{ref}}=10$) & \textbf{14.352} & \textbf{0.369} & \textbf{0.206} & \underline{0.435} & \textbf{0.686} \\
      \bottomrule
    \end{tabular}
  }
\end{table}

% \begin{figure}[t]
%     \centering
%     \includegraphics[width=1.0\linewidth]{Figs/vsvec.pdf}
%     \caption{Qualitative evaluation of vector-modality methods}. 
%     % Our method successfully synthesizes high-quality vector glyphs for complex Chinese characters in desired styles, while results generated by other existing approaches are even unrecognizable.
%     \label{vsvec_img}
% \end{figure}

\begin{table*}[t]
  \centering
  \caption{Quantitative evaluation of image-modality methods. Our method obtains performance quantitatively comparable to theirs.}
  \label{vsImg}
  \resizebox{\linewidth}{!}{
  \begin{tabular}{@{}ll|ccccc|ccccc|ccccc@{}}
    \toprule
    \multicolumn{2}{l|}{\multirow{2}{*}{Method}} & \multicolumn{5}{c|}{BTS} & \multicolumn{5}{c|}{UFTS} & \multicolumn{5}{c}{UCTS} \\
    \multicolumn{2}{l|}{} & FID \(\downarrow \) & LPIPS \(\downarrow \) & L1 \(\downarrow \) & RMSE \(\downarrow \) & SSIM \(\uparrow \) & FID \(\downarrow \) & LPIPS \(\downarrow \) & L1 \(\downarrow \) & RMSE \(\downarrow \) & SSIM \(\uparrow \) & FID \(\downarrow \) & LPIPS \(\downarrow \) & L1 \(\downarrow \) & RMSE \(\downarrow \) & SSIM \(\uparrow \) \\
    \midrule
    \multirow{4}{*}{\shortstack{Image modality\\synthesis methods}} & MX-Font & 53.564 & 0.372 & 0.204 & 0.403 & 0.567 & 78.130 & 0.383 & 0.209 & 0.410 & 0.559 & 56.427 & \underline{0.320} & \underline{0.191} & \underline{0.382} & 0.568 \\

    & CG-GAN & 65.063 & 0.457 & 0.259 & 0.464 & 0.481 & 90.584 & 0.450  & 0.250 & 0.453 & 0.478 & 58.179 & 0.424 & 0.257 & 0.457 & 0.477 \\

    & NTF & 31.406 & 0.375 & 0.203 & \underline{0.401} & 0.597 & 49.410 & \underline{0.373} & \underline{0.197} & \underline{0.396} & 0.599 & 40.348 & 0.330 & 0.197 & 0.389 & 0.579 \\

    & FontDiffuser & 10.224 & \textbf{0.259} & \textbf{0.148} & \textbf{0.310} & 0.628 & 23.378 & \textbf{0.331} & \textbf{0.179} & \textbf{0.354} & 0.592 & 15.381 & \textbf{0.294} & \textbf{0.180} & \textbf{0.352} & 0.575 \\
    \midrule
    \multirow{2}{*}{Ours} & $n_{\mathrm{ref}}=1$ & \underline{4.145} & 0.317 & 0.182 & 0.406 & \underline{0.699} & \underline{15.686} & 0.412 & 0.238 & 0.470 & \underline{0.637} & \textbf{{13.169}} & 0.381 & 0.235 & 0.469 & \underline{0.619} \\
    & $n_{\mathrm{ref}}=10$ & \textbf{3.720} & \underline{0.315} & \underline{0.180} & 0.404 & \textbf{0.701} & \textbf{14.736} & 0.403 & 0.231 & 0.463 & \textbf{0.644} & \underline{13.557} & 0.380 & 0.233 & 0.468 & \textbf{0.621} \\
    \bottomrule
  \end{tabular}
  }
\end{table*}

% \begin{figure*}[t]
%   \centering
%     \includegraphics[width=1.0\linewidth]{Figs/vsimg.pdf}
%     \caption{Qualitative evaluation of image-modality methods. We ensure high generation quality while preserving scalability. For UFTS, our model adapts to fine-grained style with only 75 glyphs per font via SFT. Representative details are highlighted in red.}
%     % \vspace{-5mm}
%     \label{vsimg_img}
% \end{figure*}

\subsection{Comparison with State-of-the-art Methods}

We compare VecFontLLM with vector-domain methods on STS, and with raster-domain methods on BTS, UFTS, and UCTS.

\paragraph{Vector-domain font synthesis methods.}
We train DeepVecFont~\cite{deepvecfont}, DeepVecFont-v2~\cite{deepvecfontv2}, and DualVector~\cite{dualvector} on 495 characters from \textbf{BFs}, matching STS.
Table~\ref{vsVec} and Fig.~\ref{vsvec_img} show that VecFontLLM clearly outperforms prior direct vector methods.
DeepVecFont and DeepVecFont-v2 often collapse into fragmented shapes.
DualVector produces denser outlines, but still distorts structure and style.
Its strong RMSE suggests that rasterized metrics can miss vector-structure errors.
Our method generates more recognizable glyphs and transfers the target style more faithfully, with further gains from $n_{\mathrm{ref}}=1$ to $n_{\mathrm{ref}}=10$.

\paragraph{Image-domain font synthesis methods.}
We compare with MX-Font~\cite{mxfont}, CG-GAN~\cite{cggan}, NTF~\cite{ntf}, and FontDiffuser~\cite{fontdiffuser} on BTS, UFTS, and UCTS.
Table~\ref{vsImg} shows that VecFontLLM achieves the best FID and SSIM on all datasets.
Although raster-domain diffusion models obtain lower LPIPS, L1, and RMSE in some cases, our method remains competitive in rendered quality and directly outputs editable vector glyphs.
As shown in Fig.~\ref{vsimg_img}, VecFontLLM preserves global shape and style comparably to image-domain methods.
On UFTS, the limited training set of 345 fonts allows both VecFontLLM and image-domain baselines to capture global style, but fine local details may still differ from the ground truth.
With lightweight supervised fine-tuning (SFT) on only 75 glyphs per font, VecFontLLM quickly adapts to the target style, as highlighted by the red boxes in Fig.~\ref{vsimg_img}.
On UCTS, the model synthesizes unseen characters by recombining learned component structures, indicating compositional generalization.

\begin{table}[t]
  \centering
  \captionof{table}{Comparison of test-time scaling strategies, showing that our proposed generation chain delivers superior results.}
  \label{tab:tsm}
  \resizebox{\linewidth}{!}{
    \begin{tabular}{@{}lll|ccccc@{}}
    \toprule
        \multicolumn{3}{l|}{\multirow{2}{*}{Method}} & \multicolumn{5}{c}{HBTS} \\
        \multicolumn{3}{l|}{} & FID \(\downarrow \) & LPIPS \(\downarrow \) & L1 \(\downarrow \) & RMSE \(\downarrow \) & SSIM \(\uparrow \) \\
    
        \midrule
        \multirow{3}{*}{\shortstack{Baseline}} & \multirow{3}{*}{Random} & random1 & 29.051 & 0.477 & 0.290 & 0.525 & 0.587 \\
        & & random2 & 29.626 & 0.478 & 0.291 & 0.526 & 0.586 \\
        & & random3 & 28.737 & 0.476 & 0.290 & 0.524 & 0.588 \\
        \midrule
        \multirow{12}{*}{\shortstack{Parallel\\thinking}} & \multirow{3}{*}{DP} & mean & 23.531 & 0.462 & 0.279 & 0.515 & 0.597 \\
        & & bottom & 23.305 & 0.461 & 0.278 & 0.514 & 0.597 \\
        & & top & 22.828 & 0.462 & 0.279 & 0.515 & 0.596 \\
    
        \cmidrule(lr){2-8}
        & \multirow{3}{*}{Entropy} & mean & 23.866 & 0.464 & 0.280 & 0.516 & 0.596 \\
        & & bottom & 23.258 & 0.461 & 0.278 & 0.514 & 0.597 \\
        & & top & 22.461 & 0.461 & 0.278 & 0.514 & 0.597 \\
        
        \cmidrule(lr){2-8}
        & \multirow{3}{*}{GI} & mean & 23.381 & 0.464 & 0.280 & 0.516 & 0.596 \\
        & & bottom & 23.312 & 0.463 & 0.280 & 0.516 & 0.596 \\
        & & top & 22.377 & 0.462 & 0.279 & 0.515 & 0.597 \\
    
        \cmidrule(lr){2-8}
        & \multirow{3}{*}{TC} & mean & \underline{21.528} & \textbf{0.455} & \textbf{0.274} & \textbf{0.510} & \textbf{0.600} \\
        & & bottom & 22.713 & 0.457 & 0.275 & 0.511 & 0.599 \\
        & & top & \textbf{20.634} & \underline{0.456} & \underline{0.275} & \underline{0.511} & \underline{0.599} \\
    
        \midrule
        \multirow{3}{*}{\shortstack{Generation\\chain}} & \multirow{3}{*}{TC} & mean & \underline{19.799} & \underline{0.453} & \underline{0.273} & \underline{0.509} & \underline{0.601} \\
        & & bottom & 23.929 & 0.461 & 0.279 & 0.515 & 0.596 \\
        & & top & \textbf{17.559} & \textbf{0.449} & \textbf{0.271} & \textbf{0.507} & \textbf{0.602} \\

        \bottomrule
    \end{tabular}  
  }
\end{table}

% \begin{figure}[t]
%   \centering
%     \includegraphics[width=\linewidth]{Figs/vsgt_dt.pdf}
%     \caption{A fine-grained comparison with ground-truth glyphs. Red boxes highlight locally consistent strokes between synthesized and ground truth.}
%     \label{vsgt_dt}
% \end{figure}

\subsection{Confidence-guided Test-time Scaling}

% We evaluate on HBTS, a hard subset of BTS with complex characters prone to Stage-1 scaffold failures.
We evaluate on HBTS, a hard subset of BTS where complex characters often cause Stage-1 scaffold failures.

\paragraph{Glyph-level selection.}
We first evaluate glyph-level selection. 
For each HBTS sample, VecFontLLM generates 20 Stage-1 scaffold candidates and selects one using different confidence scores. 
As shown in Table~\ref{tab:tsm}, confidence-based selection consistently improves over random selection. 
TC performs best among token-level measures, and different sequence-level aggregations perform similarly.

\paragraph{Component-level generation chain.}
We then evaluate the generation chain using TC as the token-level confidence measure. 
The chain samples 10 candidates for each generated component, while parallel thinking samples 20 complete scaffold candidates. 
As shown in Table~\ref{tab:tsm}, the generation chain still achieves better results. 
This suggests that component-level selection better fits long Chinese vector glyphs, as it filters uncertain structures earlier.

\subsection{Comparison with Ground Truth}

\begin{table}[t]
  \centering
  \caption{OCR accuracy comparison. }
  % \vspace{-2mm}
  \label{ocr}
  \resizebox{\columnwidth}{!}{
  \begin{tabular}{@{}l|cccc|ccc|c@{}}
    \toprule
    \multirow{2}{*}{OCR method} 
      & \multicolumn{4}{c|}{Image-modality methods [\%]} 
      & \multicolumn{3}{c|}{Ours [\%]} 
      & \multirow{2}{*}{\makecell{Ground\\truth} [\%]} \\
    & NTF & \makecell{CG-\\GAN} & \makecell{MX-\\Font} & \makecell{Font-\\Diffuser} 
      & \makecell{Direct\\generation} & \makecell{Multi-stage\\generation} & \makecell{Generation\\chain} &  \\
    \midrule
    \makecell{6763-\\classifier} & 95.14 & 97.54 & \textbf{99.72} & \underline{99.51} & 80.68 & 89.51 & 96.78 & 99.91 \\
    \makecell{Paddle\\OCR-VL} & 63.36 & 63.77 & \underline{64.61} & \textbf{65.67} & 54.10 & 59.56 & 61.82 & 70.78 \\
    \bottomrule
  \end{tabular}
  }
\end{table}

\begin{table*}
  \centering
  \caption{
  Ablation of the vector style encoder, anchor refinement, and anchor-to-curve generation
  }
  \label{tab:pipeline_ablation}
  \resizebox{\linewidth}{!}{
  \begin{tabular}{@{}ll|ccccc|ccccc|ccccc@{}}
    \toprule
    \multirow{2}{*}{Design} & \multirow{2}{*}{Variant}
    & \multicolumn{5}{c|}{BTS}
    & \multicolumn{5}{c|}{UFTS}
    & \multicolumn{5}{c}{UCTS} \\
    &
    & FID $\downarrow$ & LPIPS $\downarrow$ & L1 $\downarrow$ & RMSE $\downarrow$ & SSIM $\uparrow$
    & FID $\downarrow$ & LPIPS $\downarrow$ & L1 $\downarrow$ & RMSE $\downarrow$ & SSIM $\uparrow$
    & FID $\downarrow$ & LPIPS $\downarrow$ & L1 $\downarrow$ & RMSE $\downarrow$ & SSIM $\uparrow$ \\
    \midrule
    
    \multirow{2}{*}{\shortstack{Style\\conditioning}}
    & ViT-B/16
    & 4.157 & 0.346 & 0.196 & 0.424 & 0.683
    & 13.171 & 0.417 & 0.238 & 0.472 & 0.636
    & 12.679 & \textbf{0.380} & 0.234 & 0.469 & 0.617 \\
    & Vector style encoder
    & \textbf{3.719} & \textbf{0.324} & \textbf{0.179} & \textbf{0.404} & \textbf{0.702}
    & \textbf{12.552} & \textbf{0.399} & \textbf{0.227} & \textbf{0.460} & \textbf{0.648}
    & \textbf{12.016} & 0.386 & 0.234 & 0.469 &	\textbf{0.676} \\
    
    \midrule
    
    \multirow{2}{*}{\shortstack{Anchor\\refinement}}
    & Initial scaffold
    & 3.719 & 0.324 & 0.179 & 0.404 & 0.702
    & 12.552 & \textbf{0.399} & \textbf{0.227} & \textbf{0.460} &	\textbf{0.648}
    & 12.016 & 0.386 & 0.234 & 0.469 & \textbf{0.676} \\
    & Refined scaffold
    & \textbf{3.103} & \textbf{0.314} & \textbf{0.177} & \textbf{0.388} & \textbf{0.704}
    & \textbf{12.456} & 0.401 & 0.228 & 0.461 &	0.646
    & \textbf{10.586} & \textbf{0.376} & \textbf{0.229} & \textbf{0.453} & 0.643 \\
    
    \midrule
    
    \multirow{2}{*}{\shortstack{Anchor-to-curve\\generation}}
    & Direct generation
    & 6.271 & 0.342	& 0.197 & 0.422 & 0.686
    & 15.335 & 0.410 & 0.241 & 0.471 & 0.636
    & 14.239 & 0.395 & 0.237 & 0.473 & 0.617 \\
    & Multi-stage generation
    & \textbf{3.720} & \textbf{0.315} & \textbf{0.180} & \textbf{0.404} & \textbf{0.701} 
    & \textbf{14.736} & \textbf{0.403} & \textbf{0.231} & \textbf{0.463} & \textbf{0.644}
    & \textbf{13.557} & \textbf{0.380} & \textbf{0.233} & \textbf{0.468} & \textbf{0.621} \\
    
    \bottomrule
    \end{tabular}
  }
\end{table*}

\paragraph{Fine-grained outline structure comparison.}
Fig.~\ref{vsgt_dt} visualizes local differences between our generated outlines and the ground truth.
The highlighted regions show that our method captures local stroke curves and maintains smooth transitions between adjacent curves.
% More examples are provided in the Appendix.

\paragraph{Glyph drawing command composition comparison.}
Fig.~\ref{vsgt_in} shows the per-glyph token count distributions for four drawing elements: \texttt{<M>} (contour starts), \texttt{<L>} (line segments), \texttt{<C>} (cubic Bézier segments), and \texttt{<component>} (structural components).
The generated distributions are close to those of the ground truth.
This suggests that our outputs follow similar vector command statistics to human-designed fonts, rather than only matching rasterized appearance.

\paragraph{Optical character recognition (OCR)}
We evaluate character recognizability on BTS using a 6,763-class classifier and PaddleOCR-VL~\cite{paddleocr} (see Table~\ref{ocr}).
Image-domain methods still achieve higher OCR accuracy in several settings.
However, our vector outputs remain largely recognizable, and both the multi-stage pipeline and generation chain improve OCR accuracy.
This suggests that, together, they improve structural reliability.

\subsection{Ablation Studies}

\paragraph{Vector style encoder.}
We first ablate the vector style encoder in Stage-1 by replacing it with a pretrained ViT-B/16 encoder\cite{googlevit}.
As shown in Table~\ref{tab:pipeline_ablation}, the vector style encoder improves most metrics on three datasets, but the gains are moderate.
% This suggests that vector-domain style encoding provides useful style features.
This suggests that the encoder provides useful style cues, but style encoding alone has limited impact on vector glyph generation.

\paragraph{Anchor refinement.}
We then evaluate Stage-2 anchor refinement by comparing the refined scaffold with the initial Stage-1 scaffold.
Stage-2 improves most metrics on BTS and UCTS, but the gains on UFTS are smaller and some metrics slightly decrease.
It can refine unstable scaffold predictions, but has limited effect when Stage-1 produces structurally flawed scaffolds for unseen styles.

\paragraph{Anchor-to-curve pipeline.}
Finally, we compare direct generation with our anchor-to-curve pipeline.
The multi-stage pipeline improves all metrics across BTS, UFTS, and UCTS.
This confirms the benefit of separating anchor scaffold construction, anchor refinement, and Bézier curve completion.

Across the three ablations, the gains are smaller on UFTS and UCTS than on BTS, indicating that unseen styles and structures remain harder.
As shown in Fig.~\ref{vsimg_img}, lightweight fine-tuning with only 75 glyphs per font helps VecFontLLM quickly adapt to new styles.

\paragraph{Sampling strategy analysis.}
We study the effects of beam size, temperature, and top-\(k\) sampling.
As shown in Table~\ref{tab:gp}, a larger beam does not necessarily improve generation quality.
The best setting uses temperature \(0.8\) and top-\(k=50\), suggesting that limited sampling diversity is useful for vector glyph generation.

\begin{table}[t]
  \centering
  \captionof{table}{Parameter analysis of sampling strategies, comparing the effects of temperature, top-k, and beam size.}
  \label{tab:gp}
  \resizebox{\columnwidth}{!}{
  \begin{tabular}{@{}lll|ccccc@{}}
    \toprule
    \multicolumn{3}{l|}{\multirow{2}{*}{Method}} & \multicolumn{5}{c}{BTS} \\
    \multicolumn{3}{l|}{} & FID \(\downarrow \) & LPIPS \(\downarrow \) & L1 \(\downarrow \) & RMSE \(\downarrow \) & SSIM \(\uparrow \) \\
    
    \midrule
    \multirow{4}{*}{\shortstack{beam\_num\\=4}} & \multirow{2}{*}{\shortstack{tempera-\\ture=0.5}} & top-k=20 & 3.282 & 0.320 & 0.179 & 0.404 & \underline{0.703} \\
    & & top-k=50 & 3.720 & 0.324 & 0.180 & 0.404 & 0.701 \\
    \cmidrule(lr){2-8}
    & \multirow{2}{*}{\shortstack{tempera-\\ture=0.8}} & top-k=20 & \underline{2.896} & \underline{0.319} & \underline{0.179} & \underline{0.404} & 0.703 \\
    & & top-k=50 & \textbf{2.579} & \textbf{0.318} & \textbf{0.178} & \textbf{0.403} & \textbf{0.703} \\

    \midrule
    \multirow{4}{*}{\shortstack{beam\_num\\=16}} & \multirow{2}{*}{\shortstack{tempera-\\ture=0.5}} & top-k=20 & 4.280 & 0.322 & 0.180 & 0.405 & 0.703 \\
    & & top-k=50 & 4.193 & 0.321 & 0.180 & 0.404 & 0.703 \\
    \cmidrule(lr){2-8}
    & \multirow{2}{*}{\shortstack{tempera-\\ture=0.8}} & top-k=20 & 3.997 & 0.321 & 0.180 & 0.404 & 0.703 \\
    & & top-k=50 & 3.771 & 0.320 & 0.180 & 0.404 & 0.703 \\
    
    \bottomrule
    \end{tabular}
  }
\end{table}

\begin{table}[t]
  \centering
  \captionof{table}{Effect of supervised fine-tuning on UFTS.}
  \label{sftufts}
  \resizebox{\linewidth}{!}{
    \begin{tabular}{@{}l|ccccc@{}}
      \toprule
      \multirow{2}{*}{Method} & \multicolumn{4}{c}{UFTS} \\
      & FID $\downarrow$ & LPIPS $\downarrow$ & L1 $\downarrow$ & RMSE $\downarrow$ & SSIM $\uparrow$ \\
      \midrule
      Ours (w/o SFT) & 14.736 & 0.403 & 0.231 & 0.463 & 0.644 \\
      Ours (SFT, 75 glyphs/font) & 6.424 & 0.359 & 0.204 & 0.436 & 0.672 \\
      Ours (SFT, 775 glyphs/font) & 6.112 & 0.352 & 0.199 & 0.430 & 0.678 \\
      \bottomrule
    \end{tabular}
  }
\end{table}

% \begin{table}[t]
%   \centering
%   \caption{Inference efficiency analysis. Testing across token/glyph generation confirms our method's practical viability.}
%   % \vspace{-2mm}
%   \label{infer_time}
%   \resizebox{\columnwidth}{!}{
%   \begin{tabular}{@{}cc|cccc@{}}
%     \toprule
%     \multicolumn{2}{c|}{Token Gen. (300 tokens) [s]} & \multicolumn{4}{c}{Glyph Gen. (Avg. of 100 random glyphs) [s]} \\
%     StarCoder & Ours & Ours w/o pipeline & Ours w/ pipeline & Ours w/ parallel thinking & Ours w/ generation chain \\
    
%     \midrule
%     2.87 & 3.13 & 7.93 & 10.61 & 31.28 & 39.45 \\

%     \bottomrule
%   \end{tabular}
%   }
% \end{table}

\subsection{Experiments in Practical Applications}

% \paragraph{Inference efficiency.}
% As shown in Table~\ref{infer_time}, we evaluate inference time on an RTX 4080 GPU (FP32 precision, BS=1, Beam=4) using only K-V cache (without vLLM or FlashAttention).
% (1) Token generation: Our style/content encoder takes only 0.3s, comparable to the base LLM (StarCoder~\cite{starcoder}).
% (2) Glyph generation: Based on 100 random samples, our 3 stages pipeline is only 2.68s slower than direct generation stemming from our pipeline's ability to mitigate redundant paths.
% (1) Test-time scaling: While this method increases latency, it is practically reserved only for highly complex characters or failure recovery.

\begin{figure*}[t]
  \centering
  \begin{minipage}[t]{0.48\textwidth}
    \vspace{0pt}
    \centering
    \includegraphics[width=\linewidth]{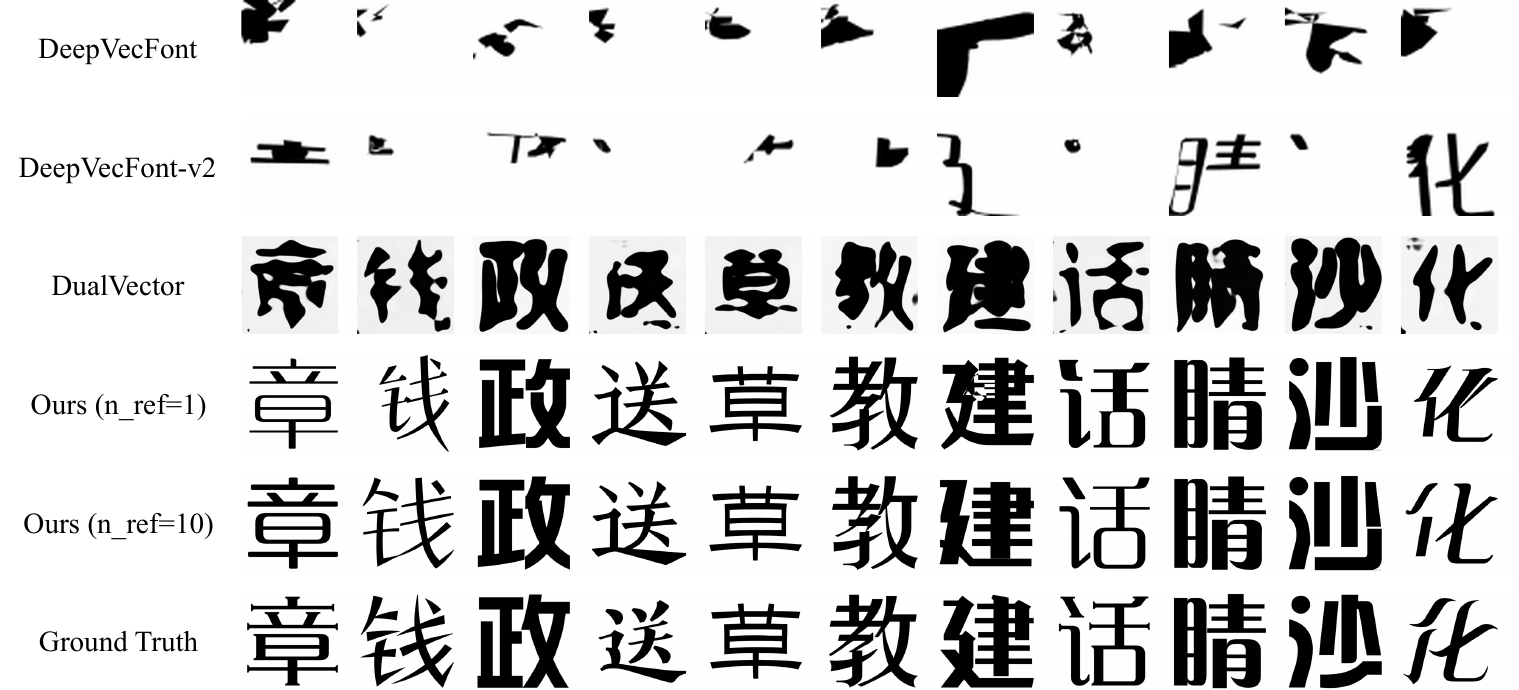}
    \caption{Qualitative evaluation of vector-modality methods.}
    \label{vsvec_img}
  \end{minipage}
  \hfill
  \begin{minipage}[t]{0.5\textwidth}
    \vspace{0pt}
    \centering
    \includegraphics[width=\linewidth]{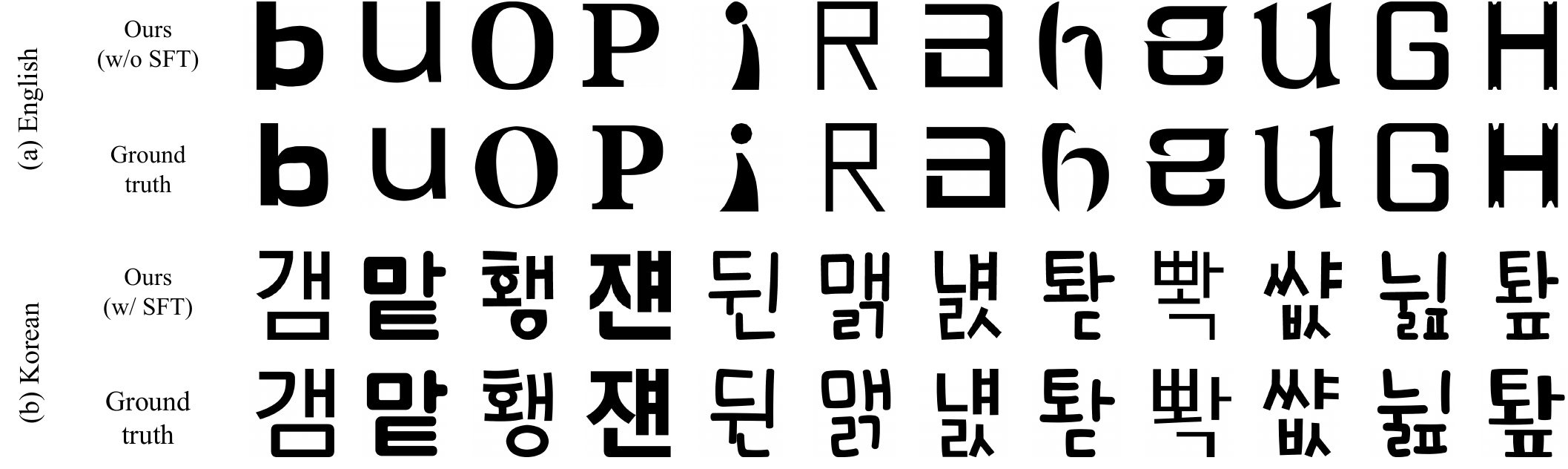}
    \caption{Unseen-font and cross-language visualization results. 
    (a) English glyph synthesis. 
    (b) Rapid adaptation to Korean under limited supervised fine-tuning.}
    \label{fig:sft_unseen}
  \end{minipage}
\end{figure*}

\begin{figure*}[t]
  \centering
    \includegraphics[width=1.0\linewidth]{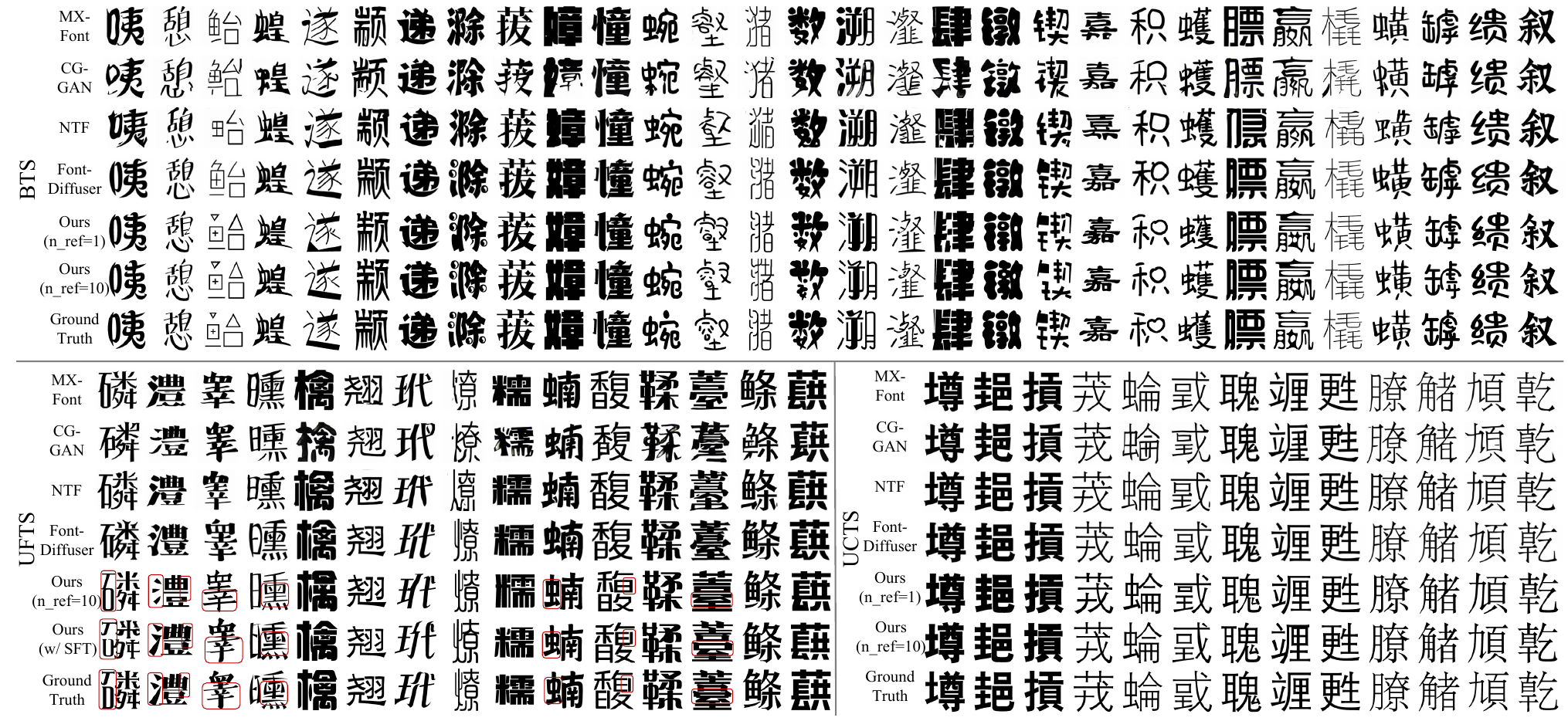}
    \caption{Qualitative evaluation of image-modality methods. We ensure high generation quality while preserving scalability. For UFTS, our model adapts to fine-grained style with only 75 glyphs per font via SFT. Representative details are highlighted in red.}
    % \vspace{-5mm}
    \label{vsimg_img}
\end{figure*}

\begin{figure*}[t]
  \centering
    \includegraphics[width=1.0\linewidth]{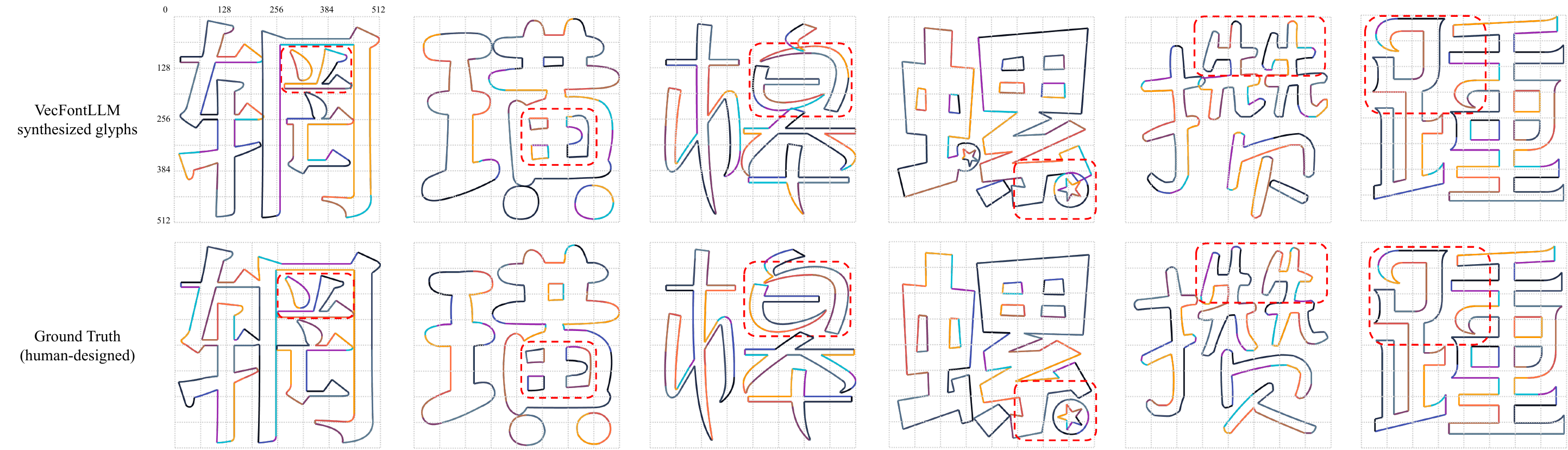}
    \caption{A fine-grained comparison with ground-truth glyphs. Red boxes highlight locally consistent strokes between the synthesized and ground truth.}
    \label{vsgt_dt}
\end{figure*}

\begin{figure*}[t]
  \centering
    \includegraphics[width=0.9\linewidth]{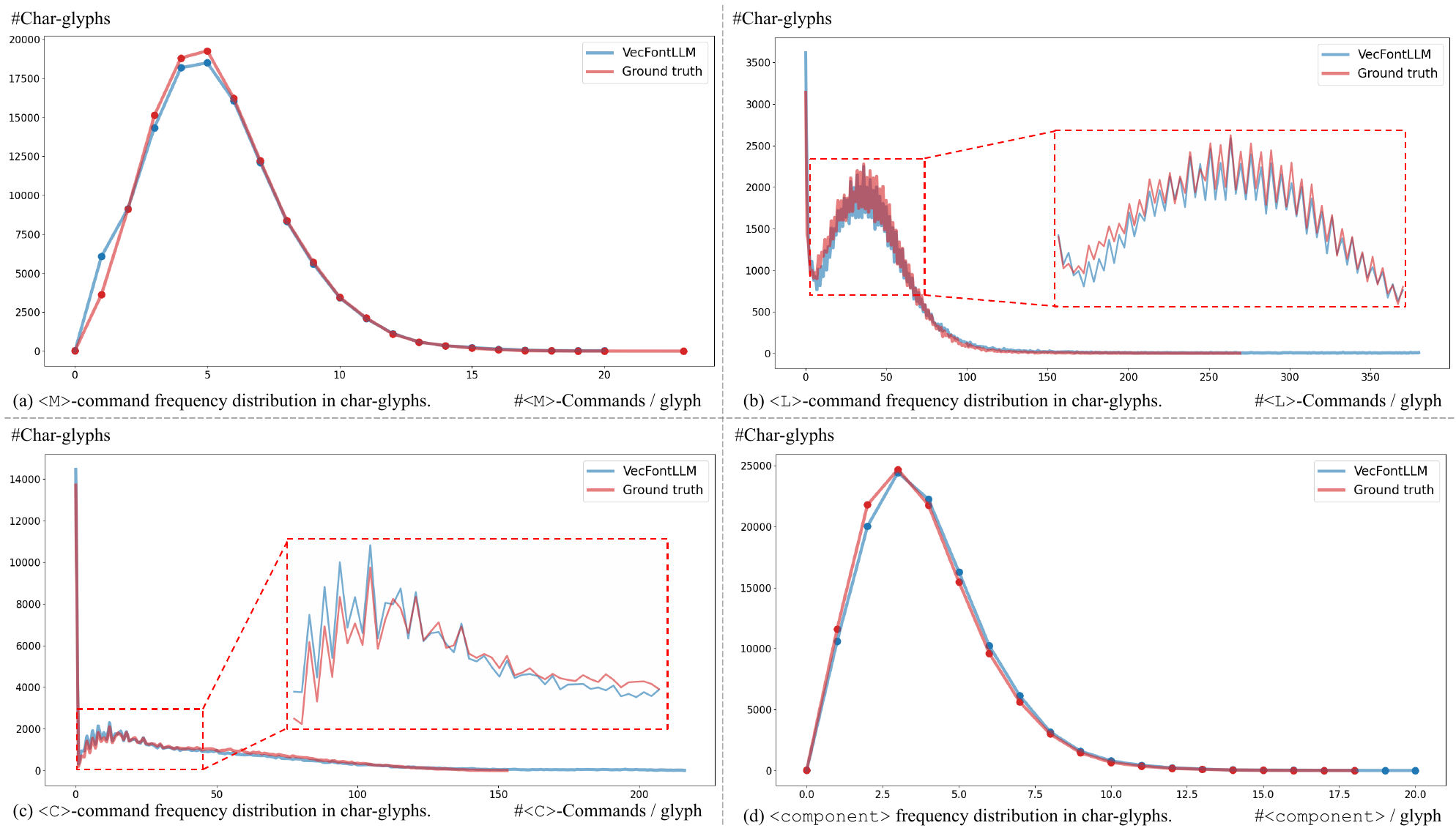}
    \caption{Token count distributions of \texttt{<M>}, \texttt{<L>}, \texttt{<C>}, and \texttt{<component>} ((a)–(d)) across vector glyphs. The generated fonts closely follow human line and curve design patterns.}
    \label{vsgt_in}
\end{figure*}

\paragraph{Multi-lingual generation.}
(1) Chinese: Beyond the 6,763 training set, our model synthesizes unseen characters by capturing structural relationships among components (UCTS of Fig.~\ref{vsimg_img}).
(2) English: By including English in progressive training, the model supports Latin scripts (Fig.~\ref{fig:sft_unseen} (a)).
(3) Korean: Despite lacking Korean in source content fonts, only Stage-1 fine-tuning (50 fonts, 2 epochs) efficiently enables Korean font generation (Fig.~\ref{fig:sft_unseen} (b)).

\paragraph{Lightweight style adaptation.}
VecFontLLM adapts to unseen font styles with lightweight supervised fine-tuning.
With only 75 glyphs per font, SFT helps the model recover local stroke terminals, stroke thickness, and global slant. 
Representative details are highlighted in Fig.~\ref{vsimg_img}.
Table~\ref{sftufts} shows clear gains across all metrics after SFT. 
Even with only 75 glyphs per font, the model substantially improves visual fidelity and reconstruction quality. 
Using 775 glyphs brings only modest additional gains, suggesting that 75-glyph SFT already approaches the performance of much larger fine-tuning sets.

\begin{figure*}
  \centering
    \includegraphics[width=0.85\linewidth]{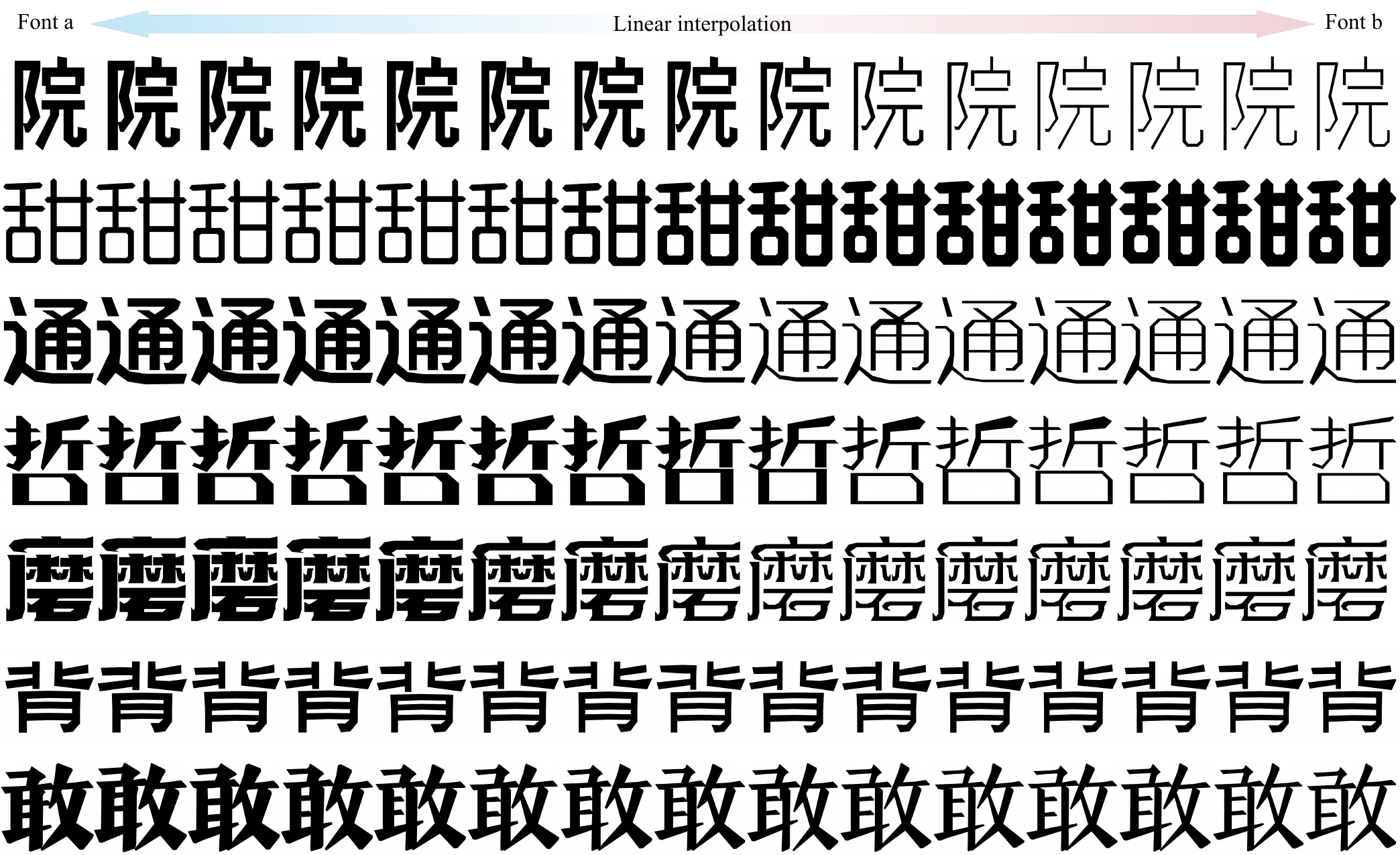}
    \caption{Interpolation results between two fonts of distinct styles. Our generative method facilitates smooth transitions between two distinct font styles, enabling progressive evolution of both glyph dimensions and stroke thickness.}
  \label{teaser5}
\end{figure*}

\paragraph{Style interpolation.}
% We linearly interpolate style features extracted by the vector style encoder and use them for generation; details are given in the Appendix.
We linearly interpolate the style features extracted by the vector style encoder and use the interpolated features for generation. 
The Appendix provides more details.
Fig.~\ref{teaser5} shows smooth interpolation between similar styles. 
The Appendix includes interpolations between more distant styles, where the results are less stable.
This is expected for vector glyphs, whose topology, command sequences, and Bézier geometry are discrete and structured.

\section{Conclusions and Limitations}

\paragraph{Conclusions.}
We introduced VecFontLLM for direct few-shot synthesis of Chinese vector fonts.
The method combines a vector-conditioned MLLM with a style encoder for vector exemplars, and an anchor-guided synthesis pipeline.
Instead of generating a single SVG sequence, VecFontLLM separates anchor scaffold construction, anchor refinement, and Bézier curve completion.
This multi-stage design decouples global structure from local curve geometry
A confidence-guided generation chain further improves scaffold generation for complex glyphs by decomposing them into component-level steps..
Experiments show clear gains over prior direct vector methods and competitive rendered quality against raster-domain baselines, while preserving editable vector outputs.

\paragraph{Limitations and future work.}
% Vector fonts remain hard to generalize and interpolate.
% Their representation is discrete and highly structured, mixing topology, commands, coordinates, and Bézier geometry in one sequence.
% This makes smooth transitions between distant styles difficult and can still lead to failures on complex glyphs.
% Our anchor scaffold reduces part of this complexity by decoupling coarse geometry from curve completion.
% Future work may further simplify vector representations around anchors or other geometric primitives to improve generalization in the native vector domain.

Vector fonts remain hard to generalize and interpolate because their discrete representation mixes topology, commands, coordinates, and Bézier geometry in one sequence.
Stage-1 instead predicts an anchor scaffold, a simpler target than the full vector outline.
Yet this scaffold is still a discrete sequence.
As a result, interpolation remains unstable between distant styles, and generalization degrade on unseen font.
Future work may explore stronger anchor generation models, such as VAE-based latent representations, to improve generalization.

% Bibliography
\clearpage
\bibliographystyle{ACM-Reference-Format}
\bibliography{sample-bibliography}

@String{Computer = "{IEEE} Computer" }

@String{Springer = "Springer-Verlag" }

@article{gan,
  title={Generative adversarial networks},
  author={Goodfellow, Ian and Pouget-Abadie, Jean and Mirza, Mehdi and Xu, Bing and Warde-Farley, David and Ozair, Sherjil and Courville, Aaron and Bengio, Yoshua},
  journal={Communications of the ACM},
  volume={63},
  number={11},
  pages={139--144},
  year={2020},
  publisher={ACM New York, NY, USA}
}

@article{diffusion,
  title={Denoising diffusion probabilistic models},
  author={Ho, Jonathan and Jain, Ajay and Abbeel, Pieter},
  journal={Advances in neural information processing systems},
  volume={33},
  pages={6840--6851},
  year={2020}
}

@inproceedings{ltm,
  title={High-resolution image synthesis with latent diffusion models},
  author={Rombach, Robin and Blattmann, Andreas and Lorenz, Dominik and Esser, Patrick and Ommer, Bj{\"o}rn},
  booktitle={Proceedings of the IEEE/CVF conference on computer vision and pattern recognition},
  pages={10684--10695},
  year={2022}
}

@article{deepsvg,
  title={Deepsvg: A hierarchical generative network for vector graphics animation},
  author={Carlier, Alexandre and Danelljan, Martin and Alahi, Alexandre and Timofte, Radu},
  journal={Advances in Neural Information Processing Systems},
  volume={33},
  pages={16351--16361},
  year={2020}
}

@article{sketchrnn,
  title={A neural representation of sketch drawings},
  author={Ha, David and Eck, Douglas},
  journal={arXiv preprint arXiv:1704.03477},
  year={2017}
}

@inproceedings{svgvae,
  title={A learned representation for scalable vector graphics},
  author={Lopes, Raphael Gontijo and Ha, David and Eck, Douglas and Shlens, Jonathon},
  booktitle={Proceedings of the IEEE/CVF International Conference on Computer Vision},
  pages={7930--7939},
  year={2019}
}

@article{deepvecfont,
  title={Deepvecfont: synthesizing high-quality vector fonts via dual-modality learning},
  author={Wang, Yizhi and Lian, Zhouhui},
  journal={ACM Transactions on Graphics (TOG)},
  volume={40},
  number={6},
  pages={1--15},
  year={2021},
  publisher={ACM New York, NY, USA}
}

@inproceedings{deepvecfontv2,
  title={Deepvecfont-v2: Exploiting transformers to synthesize vector fonts with higher quality},
  author={Wang, Yuqing and Wang, Yizhi and Yu, Longhui and Zhu, Yuesheng and Lian, Zhouhui},
  booktitle={Proceedings of the IEEE/CVF conference on computer vision and pattern recognition},
  pages={18320--18328},
  year={2023}
}

@inproceedings{vecfusion,
  title={Vecfusion: Vector font generation with diffusion},
  author={Thamizharasan, Vikas and Liu, Difan and Agarwal, Shantanu and Fisher, Matthew and Gharbi, Micha{\"e}l and Wang, Oliver and Jacobson, Alec and Kalogerakis, Evangelos},
  booktitle={Proceedings of the IEEE/CVF Conference on Computer Vision and Pattern Recognition},
  pages={7943--7952},
  year={2024}
}

@inproceedings{diffvecfont,
  title={DiffVecFont: Fusing Dual-Mode Reconstruction Vector Fonts via Masked Diffusion Transformers},
  author={Liu, Yu and Khalid, Fatimah Binti and Wang, Cunrui and Mustaffa, Mas Rina Binti and Azman, Azreen Bin},
  booktitle={International Conference on Computational Visual Media},
  pages={339--363},
  year={2025},
  organization={Springer}
}

@inproceedings{starvector,
  title={Starvector: Generating scalable vector graphics code from images and text},
  author={Rodriguez, Juan A and Puri, Abhay and Agarwal, Shubham and Laradji, Issam H and Rodriguez, Pau and Rajeswar, Sai and Vazquez, David and Pal, Christopher and Pedersoli, Marco},
  booktitle={Proceedings of the Computer Vision and Pattern Recognition Conference},
  pages={16175--16186},
  year={2025}
}

@inproceedings{chat2svg,
  title={Chat2SVG: Vector Graphics Generation with Large Language Models and Image Diffusion Models},
  author={Wu, Ronghuan and Su, Wanchao and Liao, Jing},
  booktitle={Proceedings of the Computer Vision and Pattern Recognition Conference},
  pages={23690--23700},
  year={2025}
}

@article{iconshop,
  title={Iconshop: Text-guided vector icon synthesis with autoregressive transformers},
  author={Wu, Ronghuan and Su, Wanchao and Ma, Kede and Liao, Jing},
  journal={ACM Transactions on Graphics (TOG)},
  volume={42},
  number={6},
  pages={1--14},
  year={2023},
  publisher={ACM New York, NY, USA}
}

@inproceedings{empowering,
  title={Empowering llms to understand and generate complex vector graphics},
  author={Xing, Ximing and Hu, Juncheng and Liang, Guotao and Zhang, Jing and Xu, Dong and Yu, Qian},
  booktitle={Proceedings of the Computer Vision and Pattern Recognition Conference},
  pages={19487--19497},
  year={2025}
}

@article{omnisvg,
  title={Omnisvg: A unified scalable vector graphics generation model},
  author={Yang, Yiying and Cheng, Wei and Chen, Sijin and Zeng, Xianfang and Yin, Fukun and Zhang, Jiaxu and Wang, Liao and Yu, Gang and Ma, Xingjun and Jiang, Yu-Gang},
  journal={arXiv preprint arXiv:2504.06263},
  year={2025}
}

@article{self-consistency,
  title={Self-consistency improves chain of thought reasoning in language models},
  author={Wang, Xuezhi and Wei, Jason and Schuurmans, Dale and Le, Quoc and Chi, Ed and Narang, Sharan and Chowdhery, Aakanksha and Zhou, Denny},
  journal={arXiv preprint arXiv:2203.11171},
  year={2022}
}

@article{adaptive-consistency,
  title={Let's Sample Step by Step: Adaptive-Consistency for Efficient Reasoning and Coding with LLMs},
  author={Aggarwal, Pranjal and Madaan, Aman and Yang, Yiming and others},
  journal={arXiv preprint arXiv:2305.11860},
  year={2023}
}

@article{soft-self-consistencey,
  title={Soft self-consistency improves language model agents},
  author={Wang, Han and Prasad, Archiki and Stengel-Eskin, Elias and Bansal, Mohit},
  journal={arXiv preprint arXiv:2402.13212},
  year={2024}
}

@article{fact-checking,
  title={Fact-checking the output of large language models via token-level uncertainty quantification},
  author={Fadeeva, Ekaterina and Rubashevskii, Aleksandr and Shelmanov, Artem and Petrakov, Sergey and Li, Haonan and Mubarak, Hamdy and Tsymbalov, Evgenii and Kuzmin, Gleb and Panchenko, Alexander and Baldwin, Timothy and others},
  journal={arXiv preprint arXiv:2403.04696},
  year={2024}
}

@article{self-certainty,
  title={Scalable best-of-n selection for large language models via self-certainty},
  author={Kang, Zhewei and Zhao, Xuandong and Song, Dawn},
  journal={arXiv preprint arXiv:2502.18581},
  year={2025}
}

@article{deepconf,
  title={Deep think with confidence},
  author={Fu, Yichao and Wang, Xuewei and Tian, Yuandong and Zhao, Jiawei},
  journal={arXiv preprint arXiv:2508.15260},
  year={2025}
}

@article{vae,
  title={Auto-encoding variational bayes},
  author={Kingma, Diederik P and Welling, Max},
  journal={arXiv preprint arXiv:1312.6114},
  year={2013}
}

@inproceedings{cggan,
  title={Cg-gan: An interactive evolutionary gan-based approach for facial composite generation},
  author={Zaltron, Nicola and Zurlo, Luisa and Risi, Sebastian},
  booktitle={Proceedings of the AAAI Conference on Artificial Intelligence},
  volume={34},
  number={03},
  pages={2544--2551},
  year={2020}
}

@inproceedings{mxfont,
  title={Multiple heads are better than one: Few-shot font generation with multiple localized experts},
  author={Park, Song and Chun, Sanghyuk and Cha, Junbum and Lee, Bado and Shim, Hyunjung},
  booktitle={Proceedings of the IEEE/CVF international conference on computer vision},
  pages={13900--13909},
  year={2021}
}

@inproceedings{fontdiffuser,
  title={Fontdiffuser: One-shot font generation via denoising diffusion with multi-scale content aggregation and style contrastive learning},
  author={Yang, Zhenhua and Peng, Dezhi and Kong, Yuxin and Zhang, Yuyi and Yao, Cong and Jin, Lianwen},
  booktitle={Proceedings of the AAAI conference on artificial intelligence},
  volume={38},
  number={7},
  pages={6603--6611},
  year={2024}
}

@inproceedings{ntf,
  title={Neural transformation fields for arbitrary-styled font generation},
  author={Fu, Bin and He, Junjun and Wang, Jianjun and Qiao, Yu},
  booktitle={Proceedings of the IEEE/CVF conference on computer vision and pattern recognition},
  pages={22438--22447},
  year={2023}
}

@inproceedings{zigan,
  title={Zigan: Fine-grained chinese calligraphy font generation via a few-shot style transfer approach},
  author={Wen, Qi and Li, Shuang and Han, Bingfeng and Yuan, Yi},
  booktitle={Proceedings of the 29th ACM international conference on multimedia},
  pages={621--629},
  year={2021}
}

@inproceedings{dmfont,
  title={Few-shot compositional font generation with dual memory},
  author={Cha, Junbum and Chun, Sanghyuk and Lee, Gayoung and Lee, Bado and Kim, Seonghyeon and Lee, Hwalsuk},
  booktitle={European conference on computer vision},
  pages={735--751},
  year={2020},
  organization={Springer}
}

@inproceedings{dgfont,
  title={Dg-font: Deformable generative networks for unsupervised font generation},
  author={Xie, Yangchen and Chen, Xinyuan and Sun, Li and Lu, Yue},
  booktitle={Proceedings of the IEEE/CVF conference on computer vision and pattern recognition},
  pages={5130--5140},
  year={2021}
}

@inproceedings{strokegan,
  title={Strokegan: Reducing mode collapse in chinese font generation via stroke encoding},
  author={Zeng, Jinshan and Chen, Qi and Liu, Yunxin and Wang, Mingwen and Yao, Yuan},
  booktitle={Proceedings of the AAAI conference on artificial intelligence},
  volume={35},
  number={4},
  pages={3270--3277},
  year={2021}
}

@article{difffont,
  title={Diff-font: Diffusion model for robust one-shot font generation},
  author={He, Haibin and Chen, Xinyuan and Wang, Chaoyue and Liu, Juhua and Du, Bo and Tao, Dacheng and Yu, Qiao},
  journal={International Journal of Computer Vision},
  volume={132},
  number={11},
  pages={5372--5386},
  year={2024},
  publisher={Springer}
}

@article{hfhfont,
  title={HFH-font: Few-shot Chinese font synthesis with higher quality, faster speed, and higher resolution},
  author={Li, Hua and Lian, Zhouhui},
  journal={ACM Transactions on Graphics (TOG)},
  volume={43},
  number={6},
  pages={1--16},
  year={2024},
  publisher={ACM New York, NY, USA}
}

@article{transformer,
  title={Attention is all you need},
  author={Vaswani, Ashish and Shazeer, Noam and Parmar, Niki and Uszkoreit, Jakob and Jones, Llion and Gomez, Aidan N and Kaiser, {\L}ukasz and Polosukhin, Illia},
  journal={Advances in neural information processing systems},
  volume={30},
  year={2017}
}

@article{cnn,
  title={Gradient-based learning applied to document recognition},
  author={LeCun, Yann and Bottou, L{\'e}on and Bengio, Yoshua and Haffner, Patrick},
  journal={Proceedings of the IEEE},
  volume={86},
  number={11},
  pages={2278--2324},
  year={2002},
  publisher={Ieee}
}

@article{starcoder,
  title={Starcoder: may the source be with you!},
  author={Li, Raymond and Allal, Loubna Ben and Zi, Yangtian and Muennighoff, Niklas and Kocetkov, Denis and Mou, Chenghao and Marone, Marc and Akiki, Christopher and Li, Jia and Chim, Jenny and others},
  journal={arXiv preprint arXiv:2305.06161},
  year={2023}
}

@inproceedings{dualvector,
  title={Dualvector: Unsupervised vector font synthesis with dual-part representation},
  author={Liu, Ying-Tian and Zhang, Zhifei and Guo, Yuan-Chen and Fisher, Matthew and Wang, Zhaowen and Zhang, Song-Hai},
  booktitle={Proceedings of the IEEE/CVF Conference on Computer Vision and Pattern Recognition},
  pages={14193--14202},
  year={2023}
}

@inproceedings{svgdreamer,
  title={Svgdreamer: Text guided svg generation with diffusion model},
  author={Xing, Ximing and Zhou, Haitao and Wang, Chuang and Zhang, Jing and Xu, Dong and Yu, Qian},
  booktitle={Proceedings of the IEEE/CVF Conference on Computer Vision and Pattern Recognition},
  pages={4546--4555},
  year={2024}
}

@article{text2vector,
  title={Text-to-vector generation with neural path representation},
  author={Zhang, Peiying and Zhao, Nanxuan and Liao, Jing},
  journal={ACM Transactions on Graphics (TOG)},
  volume={43},
  number={4},
  pages={1--13},
  year={2024},
  publisher={ACM New York, NY, USA}
}

@inproceedings{nivel,
  title={Nivel: Neural implicit vector layers for text-to-vector generation},
  author={Thamizharasan, Vikas and Liu, Difan and Fisher, Matthew and Zhao, Nanxuan and Kalogerakis, Evangelos and Lukac, Michal},
  booktitle={Proceedings of the IEEE/CVF Conference on Computer Vision and Pattern Recognition},
  pages={4589--4597},
  year={2024}
}

@article{sds,
  title={Dreamfusion: Text-to-3d using 2d diffusion},
  author={Poole, Ben and Jain, Ajay and Barron, Jonathan T and Mildenhall, Ben},
  journal={arXiv preprint arXiv:2209.14988},
  year={2022}
}

@article{deepseek,
  title={Deepseek-r1: Incentivizing reasoning capability in llms via reinforcement learning},
  author={Guo, Daya and Yang, Dejian and Zhang, Haowei and Song, Junxiao and Zhang, Ruoyu and Xu, Runxin and Zhu, Qihao and Ma, Shirong and Wang, Peiyi and Bi, Xiao and others},
  journal={arXiv preprint arXiv:2501.12948},
  year={2025}
}

@article{vgbench,
  title={Vgbench: Evaluating large language models on vector graphics understanding and generation},
  author={Zou, Bocheng and Cai, Mu and Zhang, Jianrui and Lee, Yong Jae},
  journal={arXiv preprint arXiv:2407.10972},
  year={2024}
}

@inproceedings{svgeditbench,
  title={SVGEditBench: A Benchmark Dataset for Quantitative Assessment of LLM's SVG Editing Capabilities},
  author={Nishina, Kunato and Matsui, Yusuke},
  booktitle={Proceedings of the IEEE/CVF Conference on Computer Vision and Pattern Recognition},
  pages={8142--8147},
  year={2024}
}

@article{reasonsvg,
  title={Reason-SVG: Hybrid Reward RL for Aha-Moments in Vector Graphics Generation},
  author={Xing, Ximing and Guan, Yandong and Zhang, Jing and Xu, Dong and Yu, Qian},
  journal={arXiv preprint arXiv:2505.24499},
  year={2025}
}

@inproceedings{svgen,
  title={SVGen: Interpretable Vector Graphics Generation with Large Language Models},
  author={Wang, Feiyu and Zhao, Zhiyuan and Liu, Yuandong and Zhang, Da and Gao, Junyu and Sun, Hao and Li, Xuelong},
  booktitle={Proceedings of the 33rd ACM International Conference on Multimedia},
  pages={9608--9617},
  year={2025}
}

@inproceedings{svgthinker,
  title={SVGThinker: Instruction-Aligned and Reasoning-Driven Text-to-SVG Generation},
  author={Chen, Hanqi and Zhao, Zhongyin and Chen, Ye and Liang, Zhujin and Ni, Bingbing},
  booktitle={Proceedings of the 33rd ACM International Conference on Multimedia},
  pages={11004--11012},
  year={2025}
}

@article{robosvg,
  title={RoboSVG: A Unified Framework for Interactive SVG Generation with Multi-modal Guidance},
  author={Wang, Jiuniu and Zhang, Gongjie and Qian, Quanhao and Gao, Junlong and Zhao, Deli and Xu, Ran},
  journal={arXiv preprint arXiv:2510.22684},
  year={2025}
}

@article{internsvg,
  title={InternSVG: Towards Unified SVG Tasks with Multimodal Large Language Models},
  author={Wang, Haomin and Yin, Jinhui and Wei, Qi and Zeng, Wenguang and Gu, Lixin and Ye, Shenglong and Gao, Zhangwei and Wang, Yaohui and Zhang, Yanting and Li, Yuanqi and others},
  journal={arXiv preprint arXiv:2510.11341},
  year={2025}
}

@article{callireader,
  title={CalliReader: Contextualizing Chinese Calligraphy via an Embedding-Aligned Vision-Language Model},
  author={Luo, Yuxuan and Tang, Jiaqi and Huang, Chenyi and Hao, Feiyang and Lian, Zhouhui},
  journal={arXiv preprint arXiv:2503.06472},
  year={2025}
}

@article{deepseekocr,
  title={DeepSeek-OCR: Contexts Optical Compression},
  author={Wei, Haoran and Sun, Yaofeng and Li, Yukun},
  journal={arXiv preprint arXiv:2510.18234},
  year={2025}
}

@article{supcon,
  title={Supervised contrastive learning},
  author={Khosla, Prannay and Teterwak, Piotr and Wang, Chen and Sarna, Aaron and Tian, Yonglong and Isola, Phillip and Maschinot, Aaron and Liu, Ce and Krishnan, Dilip},
  journal={Advances in neural information processing systems},
  volume={33},
  pages={18661--18673},
  year={2020}
}

@manual{cairo,
  title="Cairosvg",
  author="Kozea",
  note = {\url{https://cairosvg.org/}},
  year={2023}
}

@article{fid,
  title={Gans trained by a two time-scale update rule converge to a local nash equilibrium},
  author={Heusel, Martin and Ramsauer, Hubert and Unterthiner, Thomas and Nessler, Bernhard and Hochreiter, Sepp},
  journal={Advances in neural information processing systems},
  volume={30},
  year={2017}
}

@inproceedings{lpips,
  title={The unreasonable effectiveness of deep features as a perceptual metric},
  author={Zhang, Richard and Isola, Phillip and Efros, Alexei A and Shechtman, Eli and Wang, Oliver},
  booktitle={Proceedings of the IEEE conference on computer vision and pattern recognition},
  pages={586--595},
  year={2018}
}

@article{ssim,
  title={Image quality assessment: from error visibility to structural similarity},
  author={Wang, Zhou and Bovik, Alan C and Sheikh, Hamid R and Simoncelli, Eero P},
  journal={IEEE transactions on image processing},
  volume={13},
  number={4},
  pages={600--612},
  year={2004},
  publisher={IEEE}
}

@inproceedings{inceptionv3,
  title={Rethinking the inception architecture for computer vision},
  author={Szegedy, Christian and Vanhoucke, Vincent and Ioffe, Sergey and Shlens, Jon and Wojna, Zbigniew},
  booktitle={Proceedings of the IEEE conference on computer vision and pattern recognition},
  pages={2818--2826},
  year={2016}
}

@article{alexnet,
  title={Imagenet classification with deep convolutional neural networks},
  author={Krizhevsky, Alex and Sutskever, Ilya and Hinton, Geoffrey E},
  journal={Advances in neural information processing systems},
  volume={25},
  year={2012}
}

@article{rlrf4svg,
  title={Rendering-Aware Reinforcement Learning for Vector Graphics Generation},
  author={Rodriguez, Juan A and Zhang, Haotian and Puri, Abhay and Feizi, Aarash and Pramanik, Rishav and Wichmann, Pascal and Mondal, Arnab and Samsami, Mohammad Reza and Awal, Rabiul and Taslakian, Perouz and others},
  journal={arXiv preprint arXiv:2505.20793},
  year={2025}
}

@book{gi,
  title={Classification and regression trees},
  author={Breiman, Leo and Friedman, Jerome and Olshen, Richard A and Stone, Charles J},
  year={2017},
  publisher={Chapman and Hall/CRC}
}

@article{tc,
  title={Deep think with confidence},
  author={Fu, Yichao and Wang, Xuewei and Tian, Yuandong and Zhao, Jiawei},
  journal={arXiv preprint arXiv:2508.15260},
  year={2025}
}

@article{entropy,
  title={A mathematical theory of communication},
  author={Shannon, Claude E},
  journal={The Bell system technical journal},
  volume={27},
  number={3},
  pages={379--423},
  year={1948},
  publisher={Nokia Bell Labs}
}

@misc{beyondpatches,
      title={Beyond Patches: Global-aware Autoregressive Model for Multimodal Few-Shot Font Generation}, 
      author={Haonan Cai and Yuxuan Luo and Zhouhui Lian},
      year={2026},
      eprint={2601.01593},
      archivePrefix={arXiv},
      primaryClass={cs.CV},
      url={https://arxiv.org/abs/2601.01593}, 
}

@misc{vectorglyph,
      title={VecGlypher: Unified Vector Glyph Generation with Language Models}, 
      author={Xiaoke Huang and Bhavul Gauri and Kam Woh Ng and Tony Ng and Mengmeng Xu and Zhiheng Liu and Weiming Ren and Zhaochong An and Zijian Zhou and Haonan Qiu and Yuyin Zhou and Sen He and Ziheng Wang and Tao Xiang and Xiao Han},
      year={2026},
      eprint={2602.21461},
      archivePrefix={arXiv},
      primaryClass={cs.CL},
      url={https://arxiv.org/abs/2602.21461}, 
}

@article{paddleocr,
  title={Paddleocr-vl: Boosting multilingual document parsing via a 0.9 b ultra-compact vision-language model},
  author={Cui, Cheng and Sun, Ting and Liang, Suyin and Gao, Tingquan and Zhang, Zelun and Liu, Jiaxuan and Wang, Xueqing and Zhou, Changda and Liu, Hongen and Lin, Manhui and others},
  journal={arXiv preprint arXiv:2510.14528},
  year={2025}
}

@misc{googlevit,
      title={Visual Transformers: Token-based Image Representation and Processing for Computer Vision}, 
      author={Bichen Wu and Chenfeng Xu and Xiaoliang Dai and Alvin Wan and Peizhao Zhang and Zhicheng Yan and Masayoshi Tomizuka and Joseph Gonzalez and Kurt Keutzer and Peter Vajda},
      year={2020},
      eprint={2006.03677},
      archivePrefix={arXiv},
      primaryClass={cs.CV}
}

% 强制换页
\clearpage
% \begin{figure}[t]
%     \centering
%     \includegraphics[width=1.0\linewidth]{Figs/vsvec.pdf}
%     \caption{Qualitative evaluation of vector-modality methods}. 
%     \label{vsvec_img}
% \end{figure}
\clearpage
\appendix

\begin{center}
{\LARGE \textbf{Supplementary Material}}
\end{center}

\paragraph{Overview.}
This supplementary material provides further methodological insights, expanded experimental results, and an in-depth analysis of the broader applications and limitations of our approach, emphasizing its effectiveness.

The appendix is organized as follows:
\begin{itemize}
    \item Appendix~\ref{sup:cplexsvgs} describes the challenges inherent to vector representations compared with raster representations.
    \item Appendix~\ref{sup:vecrep} provides details on the handling and decoupling of vector representations.
    \item Appendix~\ref{sup:ttsm} details our test-time scaling method, including confidence metrics and the glyph synthesis algorithm.
    \item Appendix~\ref{adder} presents more visualizations of our results together with comparisons against competing methods.
    \item Appendix~\ref{sup:app} illustrates the utility of our method across various font synthesis applications.
    \item Appendix~\ref{sup:limit} discusses the limitations of our approach.
\end{itemize}

\section{Challenges of Vector Representation}
\label{sup:cplexsvgs}

\subsection{Vector Glyph Representation}

\begin{figure*}
  \centering
    \includegraphics[width=1.0\linewidth]{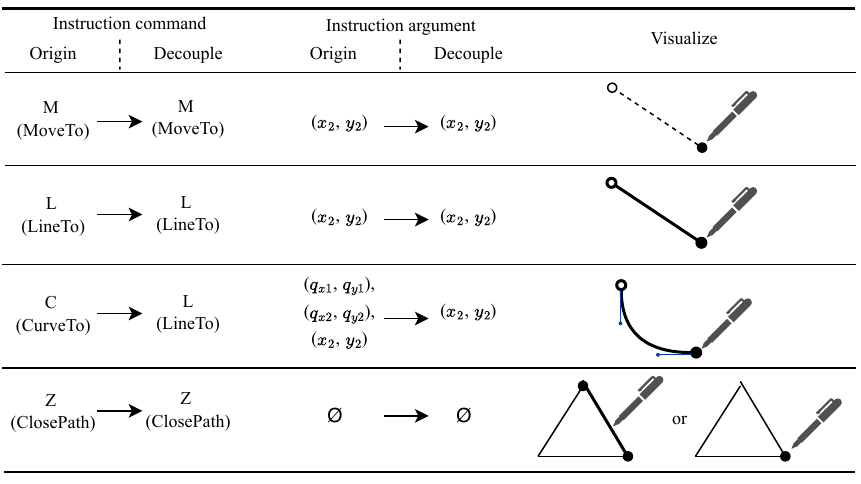}
    \caption{
    Illustration of vector glyph instructions.
    We normalize path data into four command types, \texttt{M}, \texttt{L}, \texttt{C}, and \texttt{Z}.
    Each instruction contains a command and its arguments.
    The \texttt{C} command introduces Bézier control points, while \texttt{M} and \texttt{L} only specify endpoints.
    This difference motivates our anchor-to-curve generation.
    }
  \label{vec_demo}
\end{figure*}

\begin{figure*}
  \centering
    \includegraphics[width=1.0\linewidth]{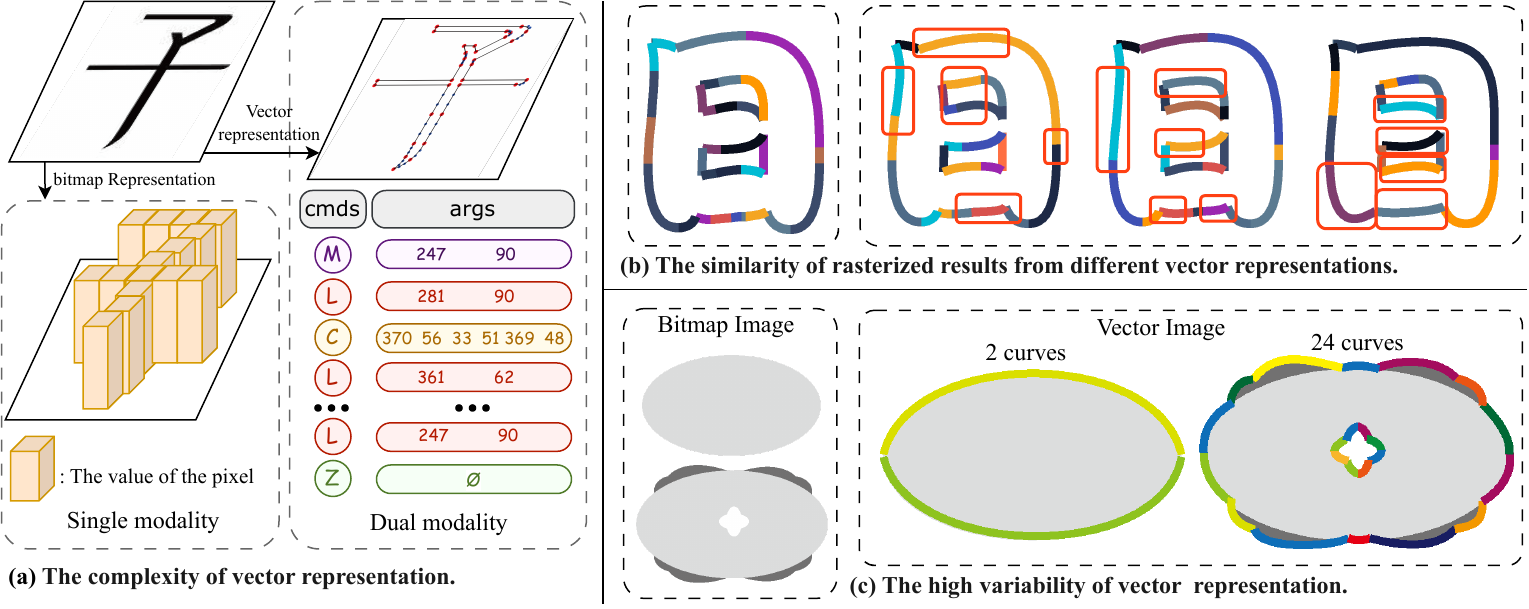}
    \caption{
    Challenges of vector generation.
    (a) Vector glyphs mix command tokens and coordinate arguments, while raster glyphs use fixed-grid pixels.
    (b) Similar raster images can have different vector command orders.
    Colored paths indicate different drawing orders.
    (c) A small visual change can require many more vector instructions.
    }
  \label{fig1}
\end{figure*}

Raster glyphs are represented by pixels on a fixed grid.
Vector glyphs are different.
They are represented by path commands, coordinates, and Bézier control points.
The model must therefore predict both the structure of the path and the geometry of each segment.

Fig.~\ref{fig1} shows three common difficulties.
First, vector glyphs mix discrete commands with continuous coordinates, as shown in Fig.~\ref{fig1}(a).
Second, the same visual shape can be encoded by different command orders or contour decompositions, as shown in Fig.~\ref{fig1}(b).
Third, a small visual change may require a large change in the vector sequence, as shown in Fig.~\ref{fig1}(c).
These properties make direct SVG generation harder than raster glyph generation.

\subsection{Complexity of Chinese Vector Fonts}

\begin{figure*}
  \centering
    \includegraphics[width=1.0\linewidth]{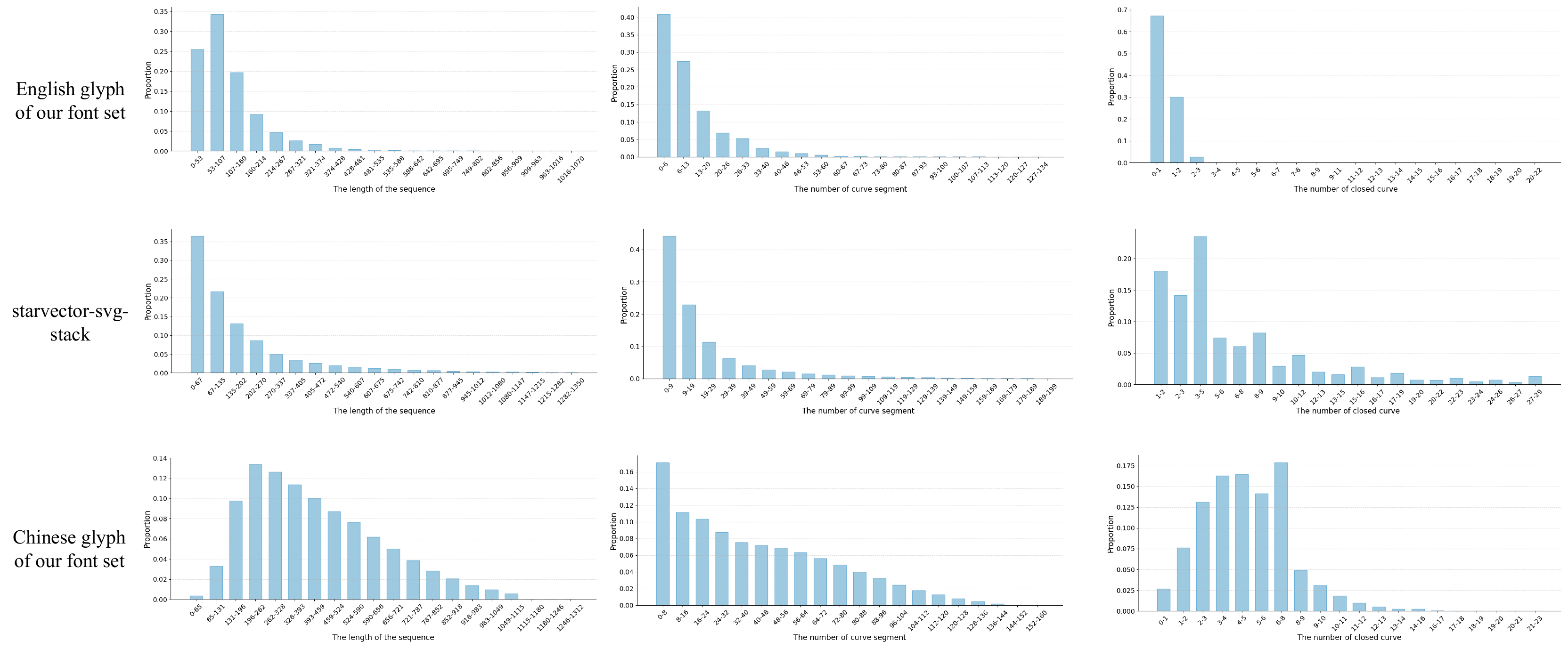}
    \caption{
    Complexity statistics of vector data.
    We compare Chinese glyphs, English glyphs from the 345 training fonts, and general vector graphics from StarVector~\cite{starvector}.
    Chinese glyphs contain longer sequences, more curve segments, and more closed paths.
    This shows that Chinese vector fonts have complex structure than English glyphs and general vector graphics.
    }
  \label{vs_eng_gra}
\end{figure*}

Chinese vector fonts further amplify these difficulties.
We compare Chinese glyphs, English glyphs from our training fonts, and general vector graphics from StarVector~\cite{starvector}.
We report three statistics: vector sequence length, the number of curve segments, and the number of closed paths.
These quantities reflect the length, curve complexity, and contour complexity of a vector glyph.

% As shown in Fig.~\ref{vs_eng_gra}, Chinese glyphs contain more long sequences, curve segments, and closed paths.
% This matches their compositional structure.
% A Chinese glyph often contains multiple components and nested contours.
% It also requires many Bézier curves to express local stroke details.
% This supports our design choice of separating anchor scaffold construction from Bézier curve completion.

As shown in Fig.~\ref{vs_eng_gra}, Chinese vector glyphs contain longer sequences, more curve segments, and more closed paths than English glyphs.
This reflects their higher compositional complexity: a Chinese character often consists of multiple components, nested contours, and dense local stroke details.
In a single vector sequence, these structural and geometric factors are predicted together, which makes direct generation.
This observation supports our decoupled design, where Stage-1 first builds an anchor scaffold for component and contour layout, and Stage-3 later completes Bézier curves for local shape and style.

\section{Handling of the Vector Representation}
\label{sup:vecrep}

\subsection{Anchor and Curve Decomposition}
\label{decvector}

We decompose each vector segment into an anchor part and a curve part.
The anchor is the endpoint of the segment.
It is used in Stage-1 and Stage-2 to construct and refine the anchor scaffold.
The curve part contains the Bézier control points.
It is predicted in Stage-3 during Bézier curve completion.

For \texttt{M} (\textit{MoveTo}) and \texttt{L} (\textit{LineTo}), each instruction has one endpoint.
We keep this endpoint as the anchor.

For \texttt{C} (\textit{CurveTo}), the last coordinate pair is the endpoint.
The two preceding coordinate pairs are Bézier control points.
We keep the endpoint in the anchor scaffold and defer the two control points to the curve completion stage.

For \texttt{Z} (\textit{ClosePath}), no coordinate is predicted.
It closes the contour.

Thus, Stage-1 and Stage-2 generate only the segment endpoints.
This gives a scaffold of the glyph.
Stage-3 then completes the missing Bézier control points to recover the final curves.

\subsection{Fixed-length Argument}
\label{unirepre}

The number of coordinates differs across commands.
To use a unified input format for the vector style encoder, we encode each instruction with four coordinate pairs:
\[
A_i = \big[(x_i^1,y_i^1), (x_i^2,y_i^2), (x_i^3,y_i^3), (x_i^4,y_i^4)\big].
\]
The first pair is the starting point of the segment.
For \(i>1\), it is copied from the endpoint of the previous instruction:
\[
(x_i^1,y_i^1) = (x_{i-1}^4,y_{i-1}^4),
\]
with \((x_0^4,y_0^4)=(0,0)\).

For \texttt{C}, the remaining three pairs define the Bézier segment.
Here \((x_i^2,y_i^2)\) and \((x_i^3,y_i^3)\) are control points, and \((x_i^4,y_i^4)\) is the endpoint.

For \texttt{M} and \texttt{L}, the endpoint is stored as \((x_i^4,y_i^4)\).
The unused control-point entries are set to zero:
\[
(x_i^2,y_i^2)=(x_i^3,y_i^3)=(0,0).
\]

For \texttt{Z}, all entries except the starting point are set to zero:
\[
(x_i^2,y_i^2)=(x_i^3,y_i^3)=(x_i^4,y_i^4)=(0,0).
\]

\section{Training Details of VecFontLLM}
\label{inittraining}

After training the vector style encoder shown in Fig.~\ref{fig2}(b), we initialize VecFontLLM with a progressive curriculum.

\begin{itemize}
    \item \textbf{Small-scale SVG pretraining.}
    We first train VecFontLLM on SVG data from 200 fonts.
    Each font contains 495 characters on a 100-pixel canvas.
    This stage adapts the backbone LLM to basic SVG font generation.

    \item \textbf{Large-scale vector-token training.}
    We then train on the full dataset described in Section~\ref{dataset}, using the final \(512\times512\) coordinate space.
    We replace raw SVG text with our discrete vector-token representation.
    Numeric coordinates are mapped to \texttt{<number>} tokens, and redundant SVG tags are removed.
    This reduces the sequence length by about \(4\times\).
\end{itemize}

After this curriculum, VecFontLLM can generate vector tokens conditioned on target content and vector style.
We then fine-tune separate stage-specific models for anchor scaffold construction, anchor refinement, and Bézier curve completion.

\section{Details of Test-time Scaling}
\label{sup:ttsm}

\subsection{Token-level Confidence Metrics}
\label{tokenlev}

For the \(i\)-th generated token, let
\[
p_i(j)=p(j\mid x,y_{<i}), \qquad j=1,\ldots,V,
\]
where \(V\) is the vocabulary size.
We compute token-level confidence from this distribution.
All metrics are defined so that a larger value indicates higher confidence.
In practice, we compute them on the sampled top-\(k\) distribution for efficiency.

\begin{itemize}
    \item \textbf{Gini purity (GP).}
    \begin{equation}
      C_i^{\mathrm{GP}}=\sum_{j=1}^{V} p_i(j)^2 .
      \label{eq:gi}
    \end{equation}

    \item \textbf{Negative entropy.}
    \begin{equation}
      C_i^{\mathrm{Ent}}=\sum_{j=1}^{V} p_i(j)\log p_i(j).
      \label{eq:entropy}
    \end{equation}

    \item \textbf{Negative distributional perplexity (DP).}
    \begin{equation}
      C_i^{\mathrm{DP}}=
      -\exp\!\left(
      -\sum_{j=1}^{V} p_i(j)\log p_i(j)
      \right).
      \label{eq:dp}
    \end{equation}

    \item \textbf{Token confidence (TC).}
    \begin{equation}
      C_i^{\mathrm{TC}}=
      -\frac{1}{V}\sum_{j=1}^{V}\log p_i(j).
      \label{eq:tc}
    \end{equation}
\end{itemize}

\subsection{Definition of \texttt{<component>}}
\label{sup:component}

\begin{figure*}
  \centering
    \includegraphics[width=1.0\linewidth]{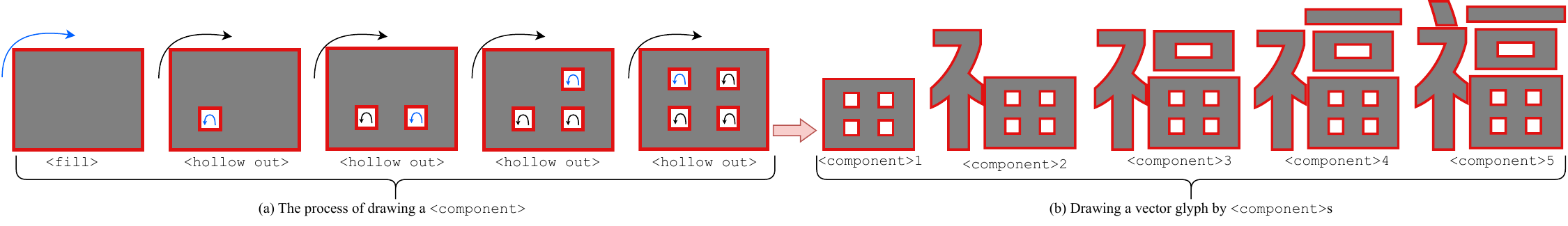}
    \caption{
    Definition of \texttt{<component>}.
    (a) A \texttt{<component>} contains one outer closed contour and its enclosed interior contours.
    (b) A vector glyph is composed of one or more \texttt{<component>}s.
    }
  \label{fig3}
\end{figure*}

Fig.~\ref{fig3}(a) illustrates the definition of \texttt{<component>}.
We use contour orientation to distinguish outer and interior contours.
A clockwise closed contour defines a filled outer region.
Counterclockwise closed contours inside it define interior holes.
We define a \texttt{<component>} as the outer contour together with all its enclosed interior contours.
It is the basic structural unit used in our generation chain.
A complete vector glyph may contain multiple \texttt{<component>}s, as shown in Fig.~\ref{fig3}(b).

\subsection{Parallel Thinking and Generation Chain}
\label{sup:ptvsgc}

\begin{figure}
  \centering
    \includegraphics[width=1.0\linewidth]{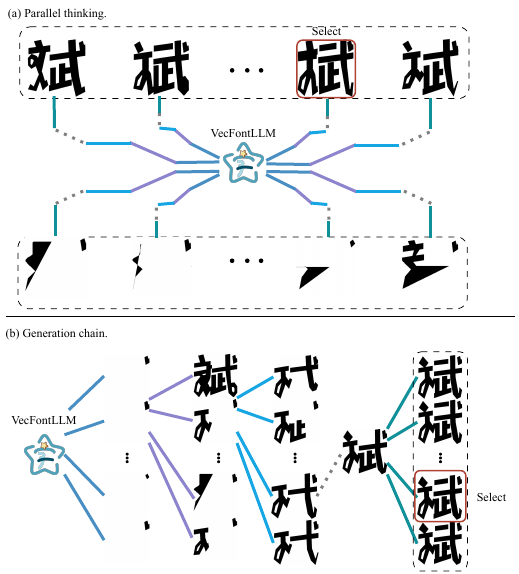}
    \caption{
    Comparison of \emph{parallel thinking} and the generation chain.
    (a) Parallel thinking samples multiple complete glyph candidates and selects the one with the highest confidence.
    (b) The generation chain samples multiple candidates for each \texttt{<component>}, appends the highest-confidence candidate to the partial scaffold, and repeats the process until the full glyph is generated.
    }
  \label{fig4}
\end{figure}

Fig.~\ref{fig4} compares the two test-time scaling strategies used in our experiments.

In \emph{parallel thinking}, the model samples multiple complete glyph candidates in one pass.
Each candidate is scored by the confidence function.
The candidate with the highest score is selected as the final output.

The generation chain applies the same confidence-based selection at the component level.
At each step, the model samples multiple \texttt{<component>} candidates conditioned on the current partial scaffold.
The highest-confidence component is appended to the scaffold before the next component is generated.
This process continues until the end-of-sequence token is produced.

\section{More Experiments and Results}
\label{adder}

\subsection{Comparison with Other Vector-domain Font Synthesis Methods}
\label{sup:vsvec}

\begin{figure*}
  \centering
    \includegraphics[width=0.85\linewidth]{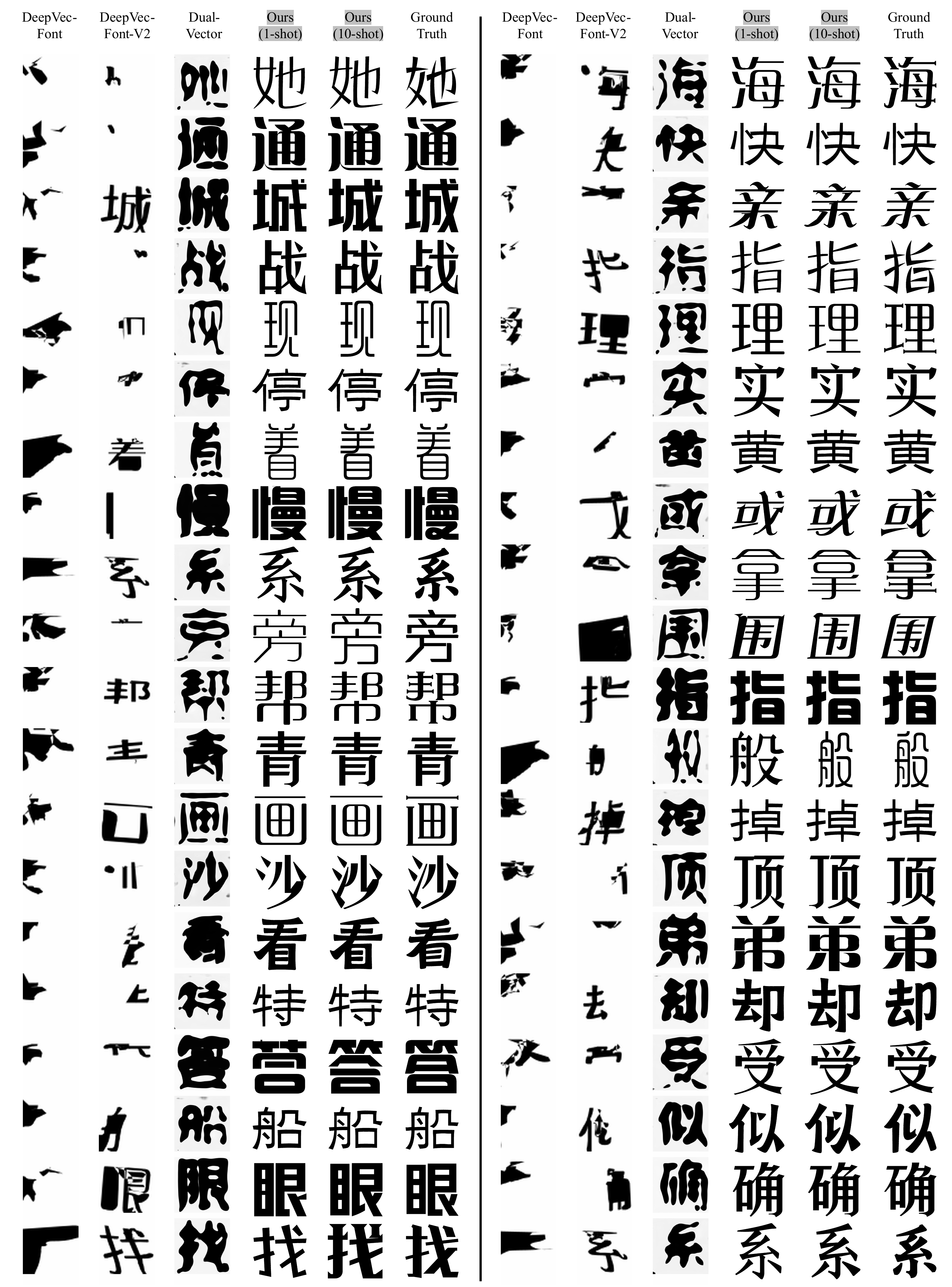}
    \caption{
    Additional visual comparison with vector-domain font synthesis methods on STS.
    }
  \label{fig:vsvec}
\end{figure*}

Fig.~\ref{fig:vsvec} shows more visual comparisons with vector-domain font synthesis methods on STS.
DeepVecFont often fails to produce recognizable Chinese glyphs.
DeepVecFont-V2 recovers partial structures, but still shows missing strokes and collapsed contours.
DualVector produces denser outlines, but the character structure and style often deviate from the reference.
VecFontLLM generates more legible glyphs and better preserves the target style.

\subsection{Comparison with Other Image-domain Font Synthesis Methods}
\label{sup:vsbit}

\begin{figure*}
  \centering
    \includegraphics[width=0.8\linewidth]{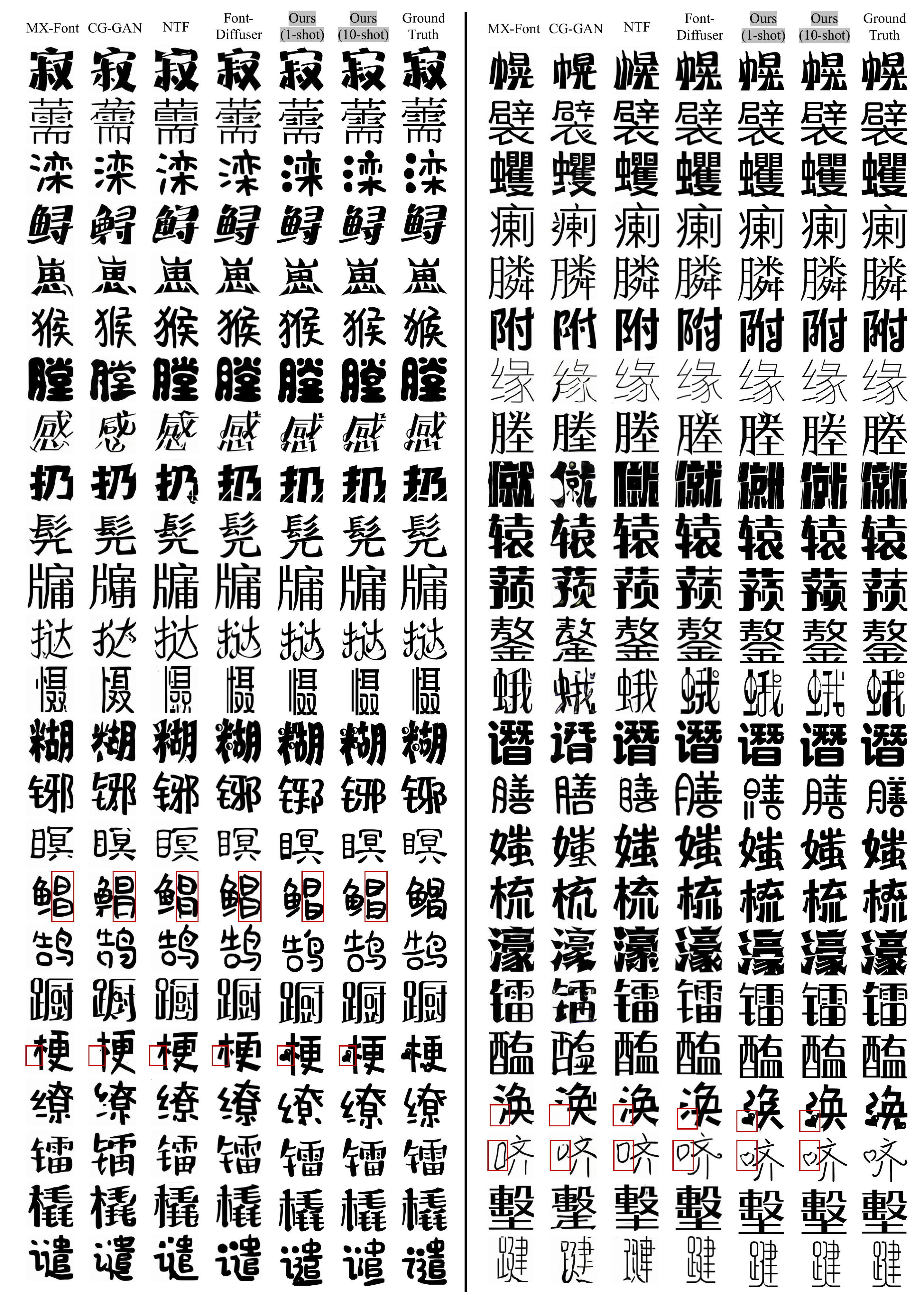}
    \caption{
    Additional visual comparison with image-domain font synthesis methods on BTS.
    }
  \label{fig:vsbit_1}
\end{figure*}

\begin{figure*}
  \centering
    \includegraphics[width=0.8\linewidth]{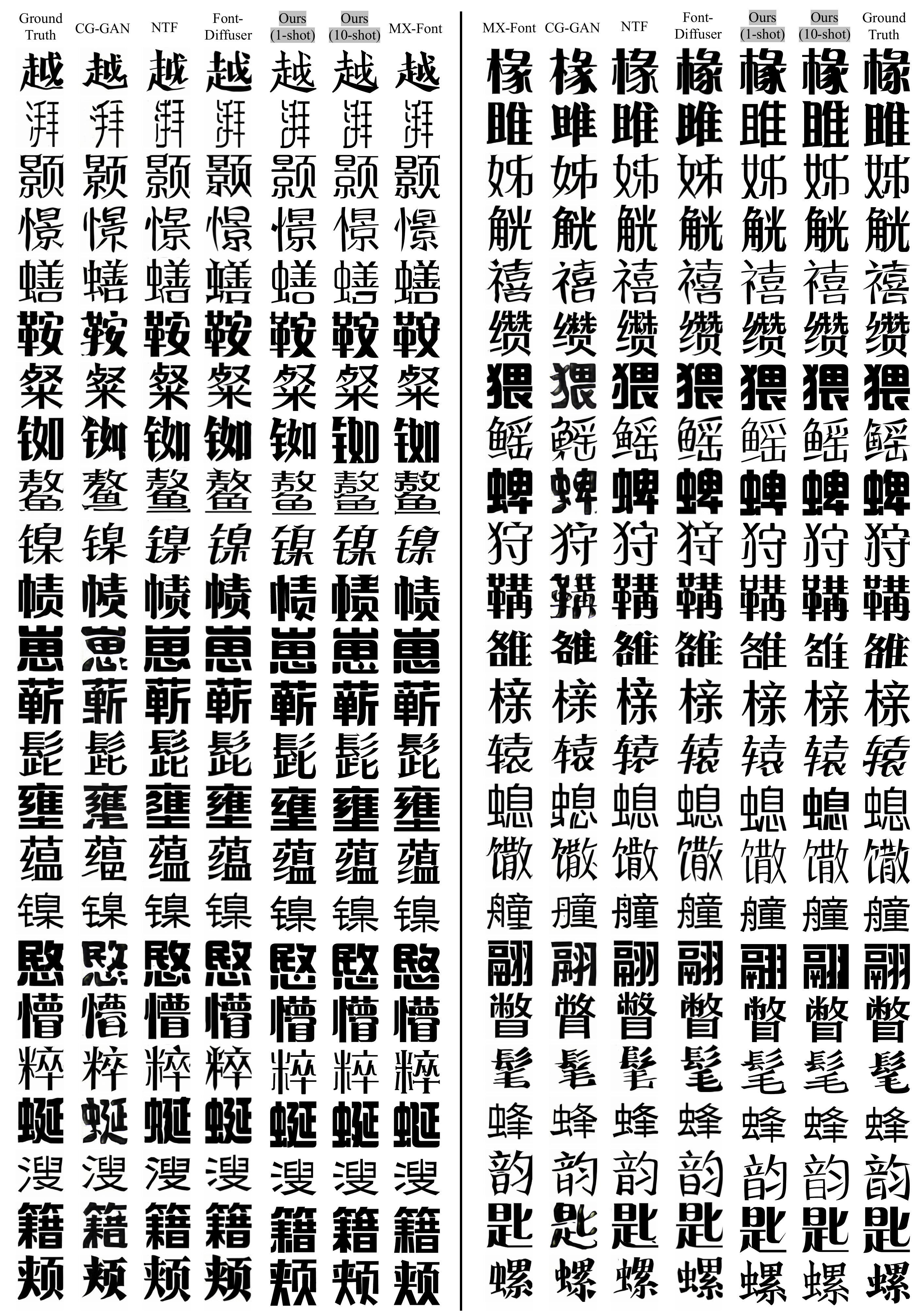}
    \caption{
    Additional visual comparison with image-domain font synthesis methods on \textbf{UFTS}.
    }
  \label{fig:vsbit_2}
\end{figure*}

\begin{figure*}
  \centering
    \includegraphics[width=0.8\linewidth]{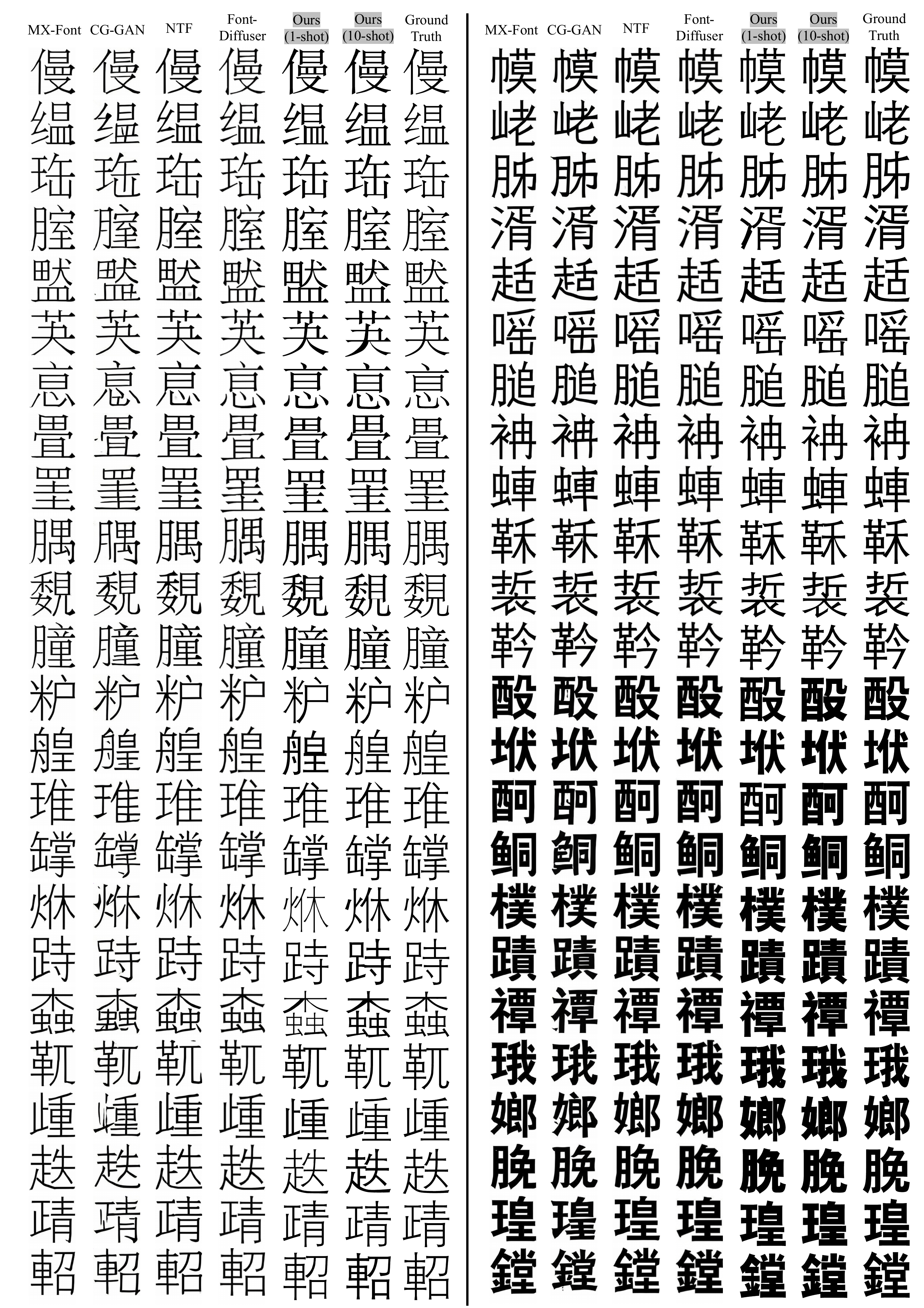}
    \caption{
    Additional visual comparison with image-domain font synthesis methods on \textbf{UCTS}.
    }
  \label{fig:vsbit_3}
\end{figure*}

Fig.~\ref{fig:vsbit_1}, Fig.~\ref{fig:vsbit_2}, and Fig.~\ref{fig:vsbit_3} show additional comparisons on BTS, \textbf{UFTS}, and \textbf{UCTS}.
On BTS, several image-domain methods show local artifacts, such as irregular curves or inaccurate stroke details.
The red boxes in Fig.~\ref{fig:vsbit_1} highlight these differences.
VecFontLLM produces smoother local curves while keeping editable vector outputs.

On \textbf{UFTS}, both VecFontLLM and image-domain methods capture the global style of unseen fonts.
Fine local details may still differ from the ground truth, due to the limited training set of 345 fonts.
On \textbf{UCTS}, VecFontLLM remains comparable to image-domain methods on unseen characters.
Failures still occur on highly complex glyphs, especially when local stroke structures are dense.

\subsection{Qualitative Comparisons with Ground Truth}
\label{sup:vsgt}

\begin{figure*}
  \centering
    \includegraphics[width=0.75\linewidth]{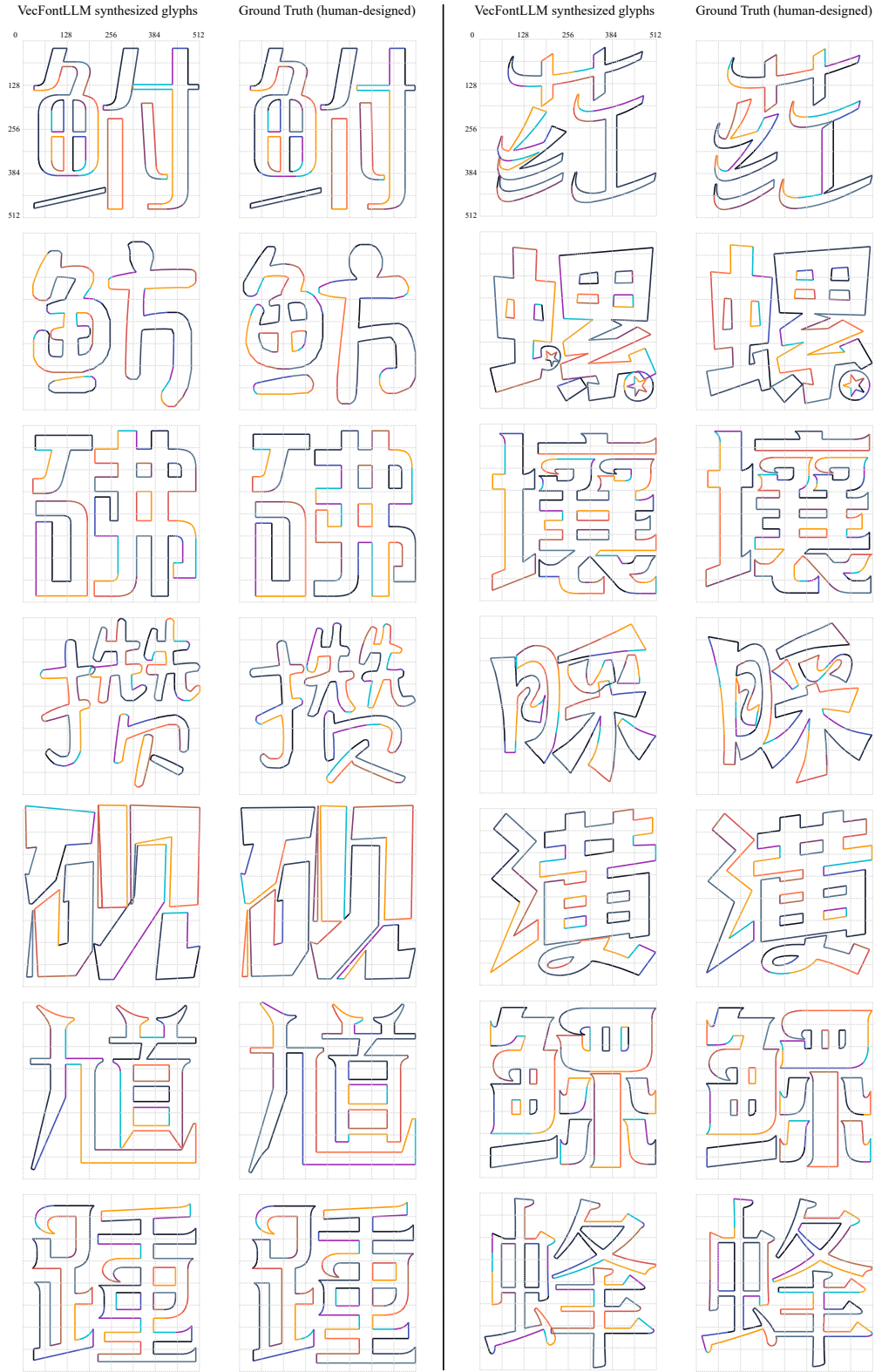}
    \caption{
    Additional comparisons between our generated vector glyphs and the ground truth.
    The results show smooth local curves and consistent outline structures.
    }
  \label{fig:vsgt_dt_}
\end{figure*}

Fig.~\ref{fig:vsgt_dt_} shows more comparisons with ground-truth vector glyphs.
Our results preserve many local curve details and maintain smooth transitions between adjacent Bézier segments.
The global arrangement of lines and curves is also close to the ground truth in most cases.
These results suggest that Stage-3 Bézier curve completion can recover local curvature while keeping the refined anchor scaffold stable.

\subsection{Confidence-guided Test-time Scaling}
\label{sup:ttsme}

\begin{figure*}
  \centering
    \includegraphics[width=0.7\linewidth]{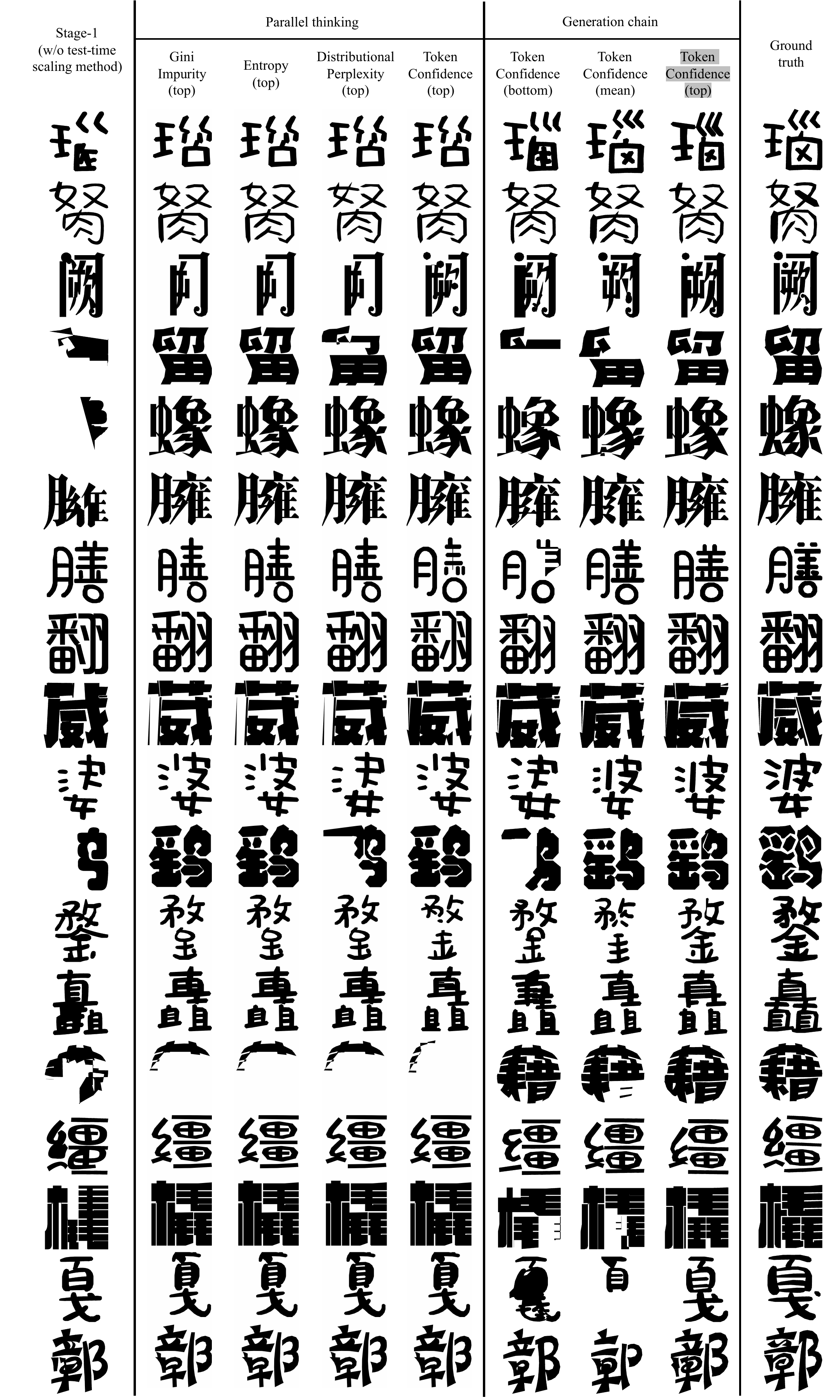}
    \caption{
    Additional comparison of test-time scaling strategies on HBTS.
    The generation chain with token confidence and top-45\% aggregation gives the best overall results.
    }
  \label{fig:tts}
\end{figure*}

Fig.~\ref{fig:tts} compares different test-time scaling strategies during Stage-1 scaffold generation on HBTS.
The generation chain with token confidence and top-45\% sequence aggregation gives the best overall visual quality.
It better preserves the scaffold structure and reduces local failures.
Since sampling is stochastic, other strategies may still produce better results for some individual glyphs.

\subsection{Ablation Studies}
\label{sup:abl}

\begin{figure*}
  \centering
    \includegraphics[width=0.73\linewidth]{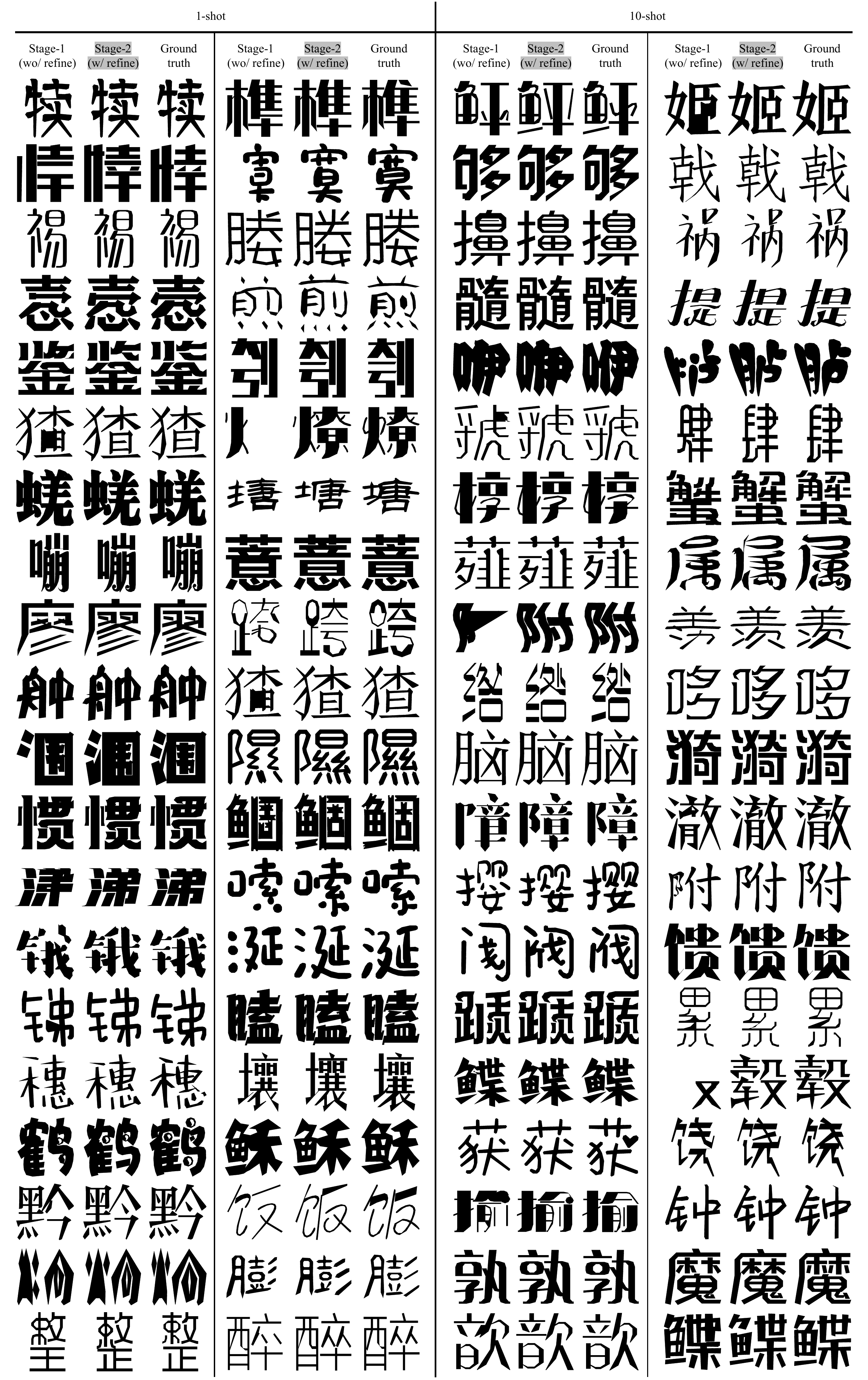}
    \caption{
    Qualitative ablation of Stage-2 anchor refinement.
    Stage-2 corrects missing parts and structural errors in the initial anchor scaffold.
    }
  \label{fig:ablr}
\end{figure*}

\paragraph{Anchor refinement.}
We compare the initial Stage-1 scaffold with the refined Stage-2 output on BTS.
As shown in Fig.~\ref{fig:ablr}, Stage-1 may produce collapsed shapes, missing parts, or incorrect local structures.
Stage-2 corrects many of these errors by refining the anchor scaffold.
It does not improve every glyph, but the quantitative results in Table~\ref{tab:pipeline_ablation} show a clear overall gain.

\begin{figure*}
  \centering
    \includegraphics[width=0.73\linewidth]{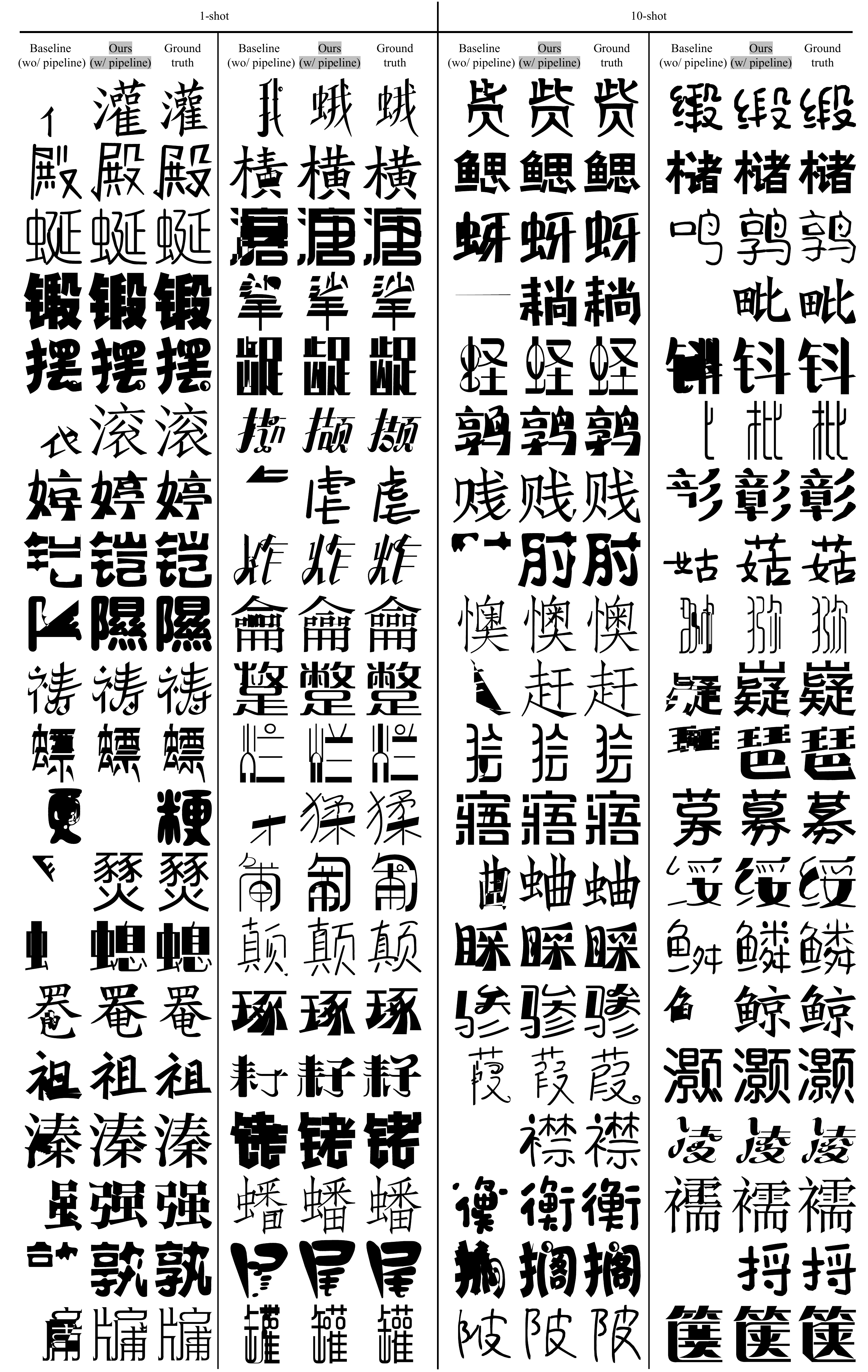}
    \caption{
    Qualitative ablation of the anchor-to-curve pipeline.
    The multi-stage pipeline reduces missing parts, distorted structures, and generation failures.
    }
  \label{fig:ablp}
\end{figure*}

\paragraph{Anchor-to-curve pipeline.}
We use the progressively initialized VecFontLLM without stage-specific pipeline training as the baseline.
We compare it with the full multi-stage pipeline.
As shown in Fig.~\ref{fig:ablp}, the baseline often produces missing parts, collapsed outlines, or invalid character structures.
The multi-stage pipeline reduces these failures by separating anchor scaffold construction, anchor refinement, and Bézier curve completion.
The quantitative results in Table~\ref{tab:pipeline_ablation} show the same trend.

\paragraph{Sampling strategy analysis.}

\begin{figure*}
  \centering
    \includegraphics[width=0.90\linewidth]{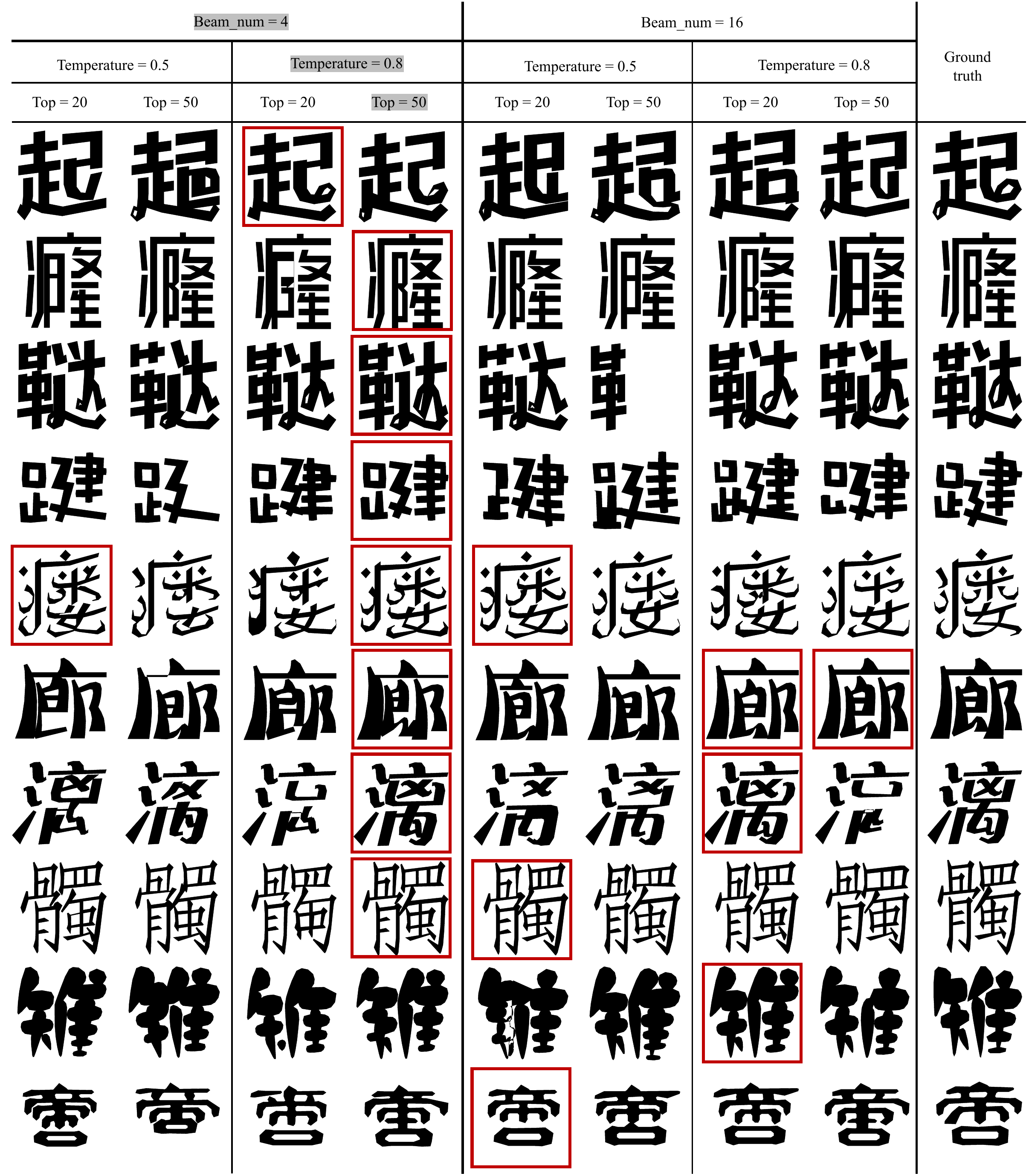}
    \caption{
    Qualitative comparison under different sampling parameters.
    The red box marks the result closest to the ground truth.
    }
  \label{fig:comp}
\end{figure*}

Fig.~\ref{fig:comp} shows examples generated with different sampling parameters.
The red box marks the result closest to the ground truth.
The visually better results are often consistent with the best quantitative settings in Table~\ref{tab:gp}.
Still, the trend is not deterministic.
Good samples can appear under other settings, which reflects the stochastic nature of autoregressive vector generation.

\subsection{Vector Font Interpolation}

\begin{figure*}
  \centering
    \includegraphics[width=1.0\linewidth]{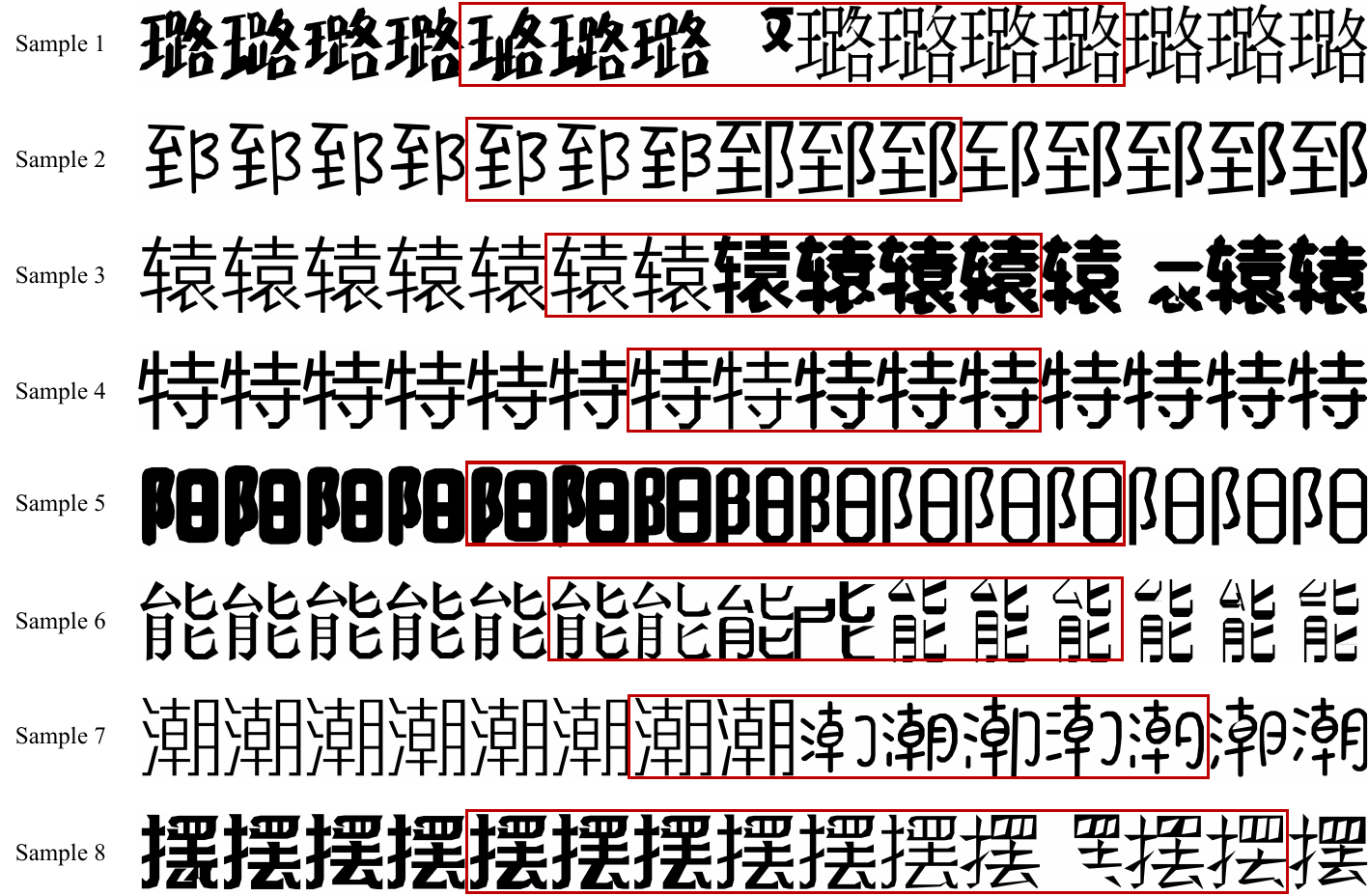}
    \caption{
    Interpolation between vector fonts.
    The red boxes highlight the gradual change of local geometric features.
    }
  \label{fig:fontinterp}
\end{figure*}

We linearly interpolate two style features extracted by the vector style encoder:
\begin{equation}
  h_{\mathrm{interp}}^s = (1-\alpha) h_a^s + \alpha h_b^s,
  \qquad \alpha \in [0,1].
  \label{fontinterp}
\end{equation}
The interpolated feature is then used as the style condition for generation.

Fig.~\ref{fig:fontinterp} shows the interpolation results.
For similar font pairs, such as samples 4, 5, and 8, the transition is smooth.
Stroke width and local handwriting traits change gradually.
For more distant font pairs, the transition becomes less stable.
Samples 2, 3, and 7 show abrupt style changes, while samples 1 and 6 show generation failures.
This reflects the difficulty of interpolating vector fonts.
Their style is tied to discrete topology, command sequences, and Bézier geometry.

\begin{table*}[t]
  \centering
  \caption{Inference efficiency analysis. Testing across token/glyph generation confirms our method's practical viability.}
  \label{infer_time}
  \resizebox{\textwidth}{!}{
  \begin{tabular}{@{}cc|cccc@{}}
    \toprule
    \multicolumn{2}{c|}{Token Gen. (300 tokens) [s]} & \multicolumn{4}{c}{Glyph Gen. (Avg. of 100 random glyphs) [s]} \\
    backbone LLM & Ours & Ours w/o pipeline & Ours w/ pipeline & Ours w/ parallel thinking & Ours w/ generation chain \\
    
    \midrule
    2.87 & 3.13 & 7.93 & 10.61 & 31.28 & 39.45 \\

    \bottomrule
  \end{tabular}
  }
\end{table*}

\subsection{Inference efficiency}
As shown in Table~\ref{infer_time}, we evaluate inference time on an RTX 4080 GPU (FP32 precision, BS=1, Beam=4) using only K-V cache (without vLLM or FlashAttention).
(1) Token generation: Our style/content encoder takes only 0.3s, comparable to the backbone LLM (StarCoder~\cite{starcoder}).
(2) Glyph generation: Based on 100 random samples, our multi-stage pipeline is only 2.68s slower than direct generation steming from our pipeline's ability to mitigate redundant paths.
(1) Test-time scaling: While this method increases latency, it is practically reserved only for highly complex characters or failure recovery.

\begin{figure}
  \centering
    \includegraphics[width=1.0\linewidth]{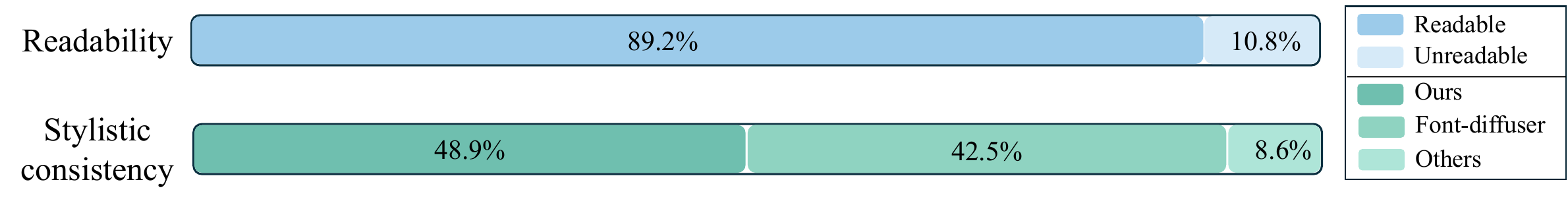}
    \caption{Human evaluation: Our approach yields robust readability and the highest preference rate for stylistic consistency.}
  \label{humaneva}
\end{figure}

\subsection{Human evaluation}
We conduct user studies on readability and stylistic consistency.
For readability, 89.2\% of our generated glyphs are rated readable over 50 test glyphs.
For stylistic consistency, participants compare 20 generation tasks and choose the result closest to the target font style.
Our method obtains the highest preference rate, with 48.9\% of votes, compared with 42.5\% for FontDiffuser and 8.6\% for other methods (Fig.~\ref{humaneva}).
These results support the perceptual quality of our vector glyphs.

\section{Applications}
\label{sup:app}

\paragraph{Direct vector glyph synthesis.}
Fig.~\ref{fig:vsbit_1}, Fig.~\ref{fig:vsbit_2}, and Fig.~\ref{fig:vsbit_3} show direct vector glyph synthesis results.
In this setting, a raster glyph provides the target content, and several vector glyphs provide the style reference.
The model directly outputs editable vector glyphs.

\begin{figure}
  \centering
    \includegraphics[width=0.95\linewidth]{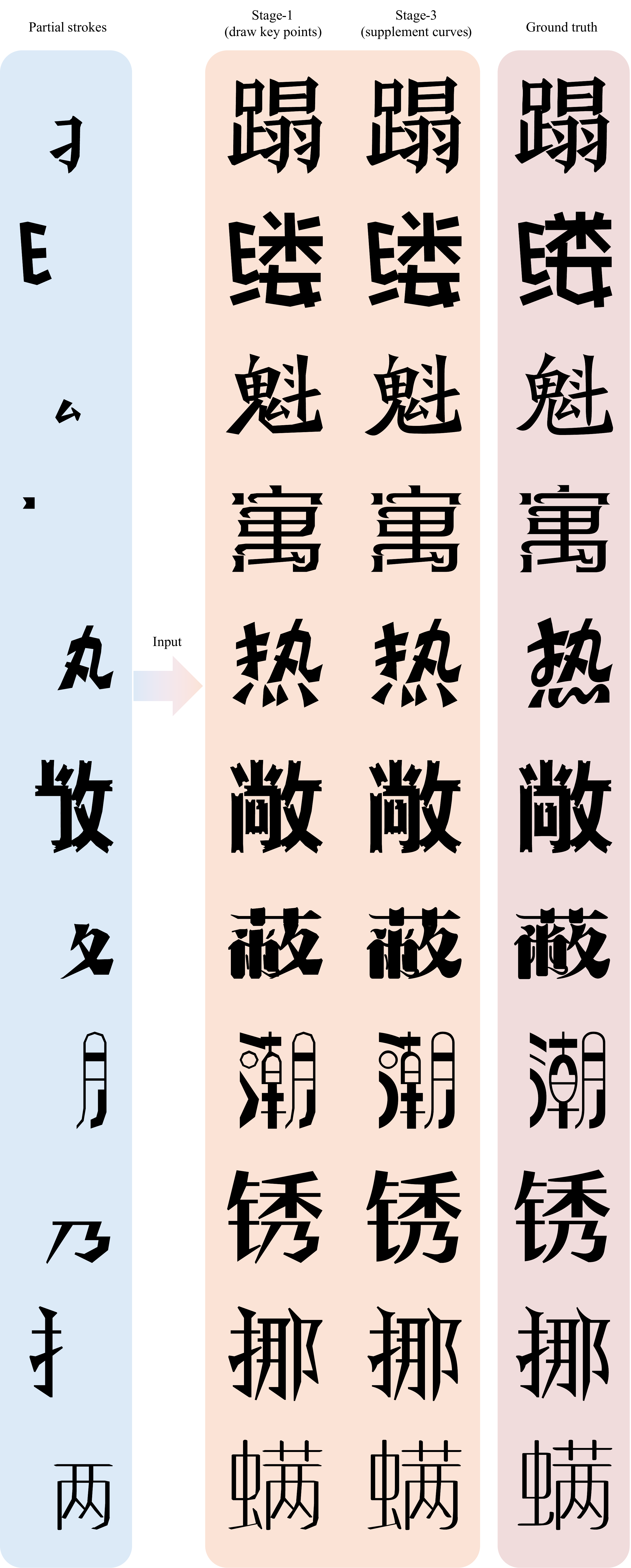}
    \caption{
    Qualitative results of vector glyph completion.
    Given a partial vector glyph, our method completes the missing structure while preserving the provided input.
    }
  \label{fig:compg}
\end{figure}

\paragraph{Vector glyph completion.}
The autoregressive design also supports completion from a partial vector glyph.
As shown in Fig.~\ref{fig:compg}, the user-provided partial structure is used as a condition for generation.
We use Stage-1 and Stage-3 in this setting, and skip Stage-2 to avoid changing the input scaffold.
This allows users to keep designed parts of a glyph and synthesize the missing vector geometry.

\section{Limitations}
\label{sup:limit}

\begin{figure}
  \centering
    \includegraphics[width=0.88\linewidth]{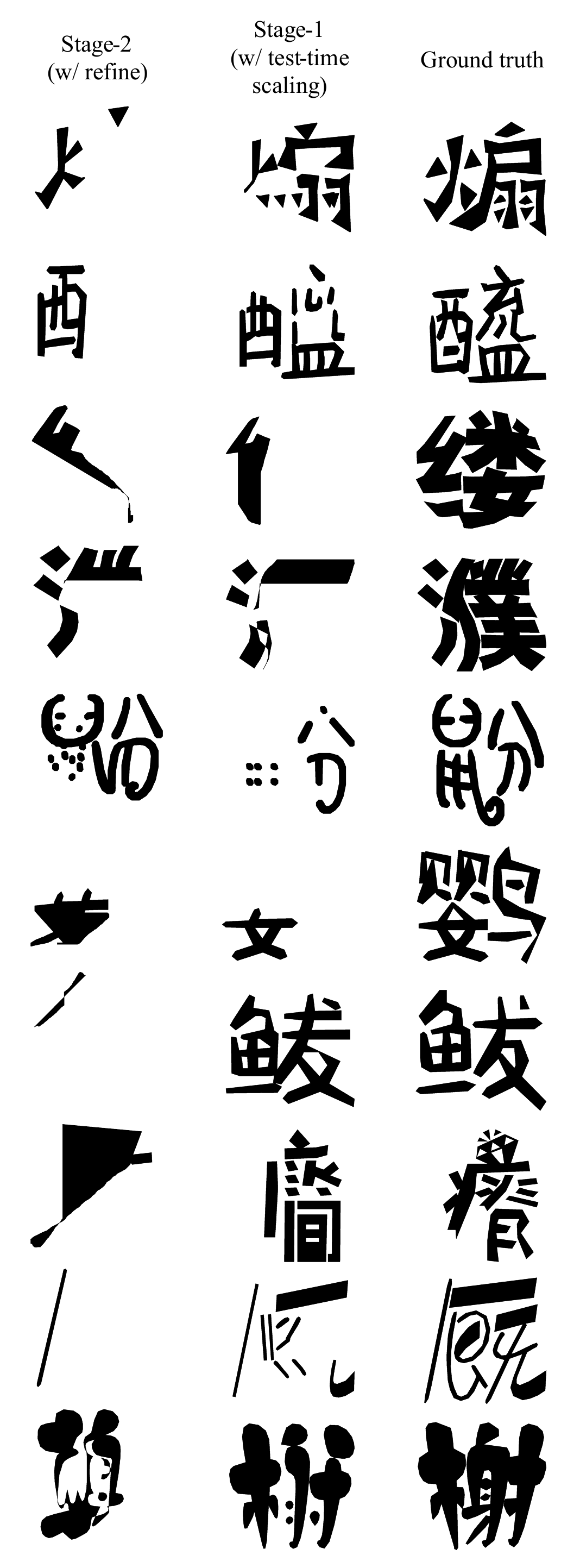}
    \caption{Examples of generation failure. We generate only a partial character structure, accompanied by meaningless vector sequences that, when rasterized, produce nonsensical images.}
  \label{fig:genf}
\end{figure}

\paragraph{Generation failures.}
Our font synthesis pipeline and test-time scaling method substantially reduce generation failures, but they still cannot fully eliminate them. 
As shown in Fig.~\ref{fig:genf}, failures persist when synthesizing highly complex glyphs. 
These cases arise from the model’s difficulty in learning stable token-distribution patterns for intricate glyphs. 
Deviations from the learned distribution during sampling can push the model into a collapse state, producing invalid or meaningless token sequences.

\begin{figure}
  \centering
    \includegraphics[width=0.89\linewidth]{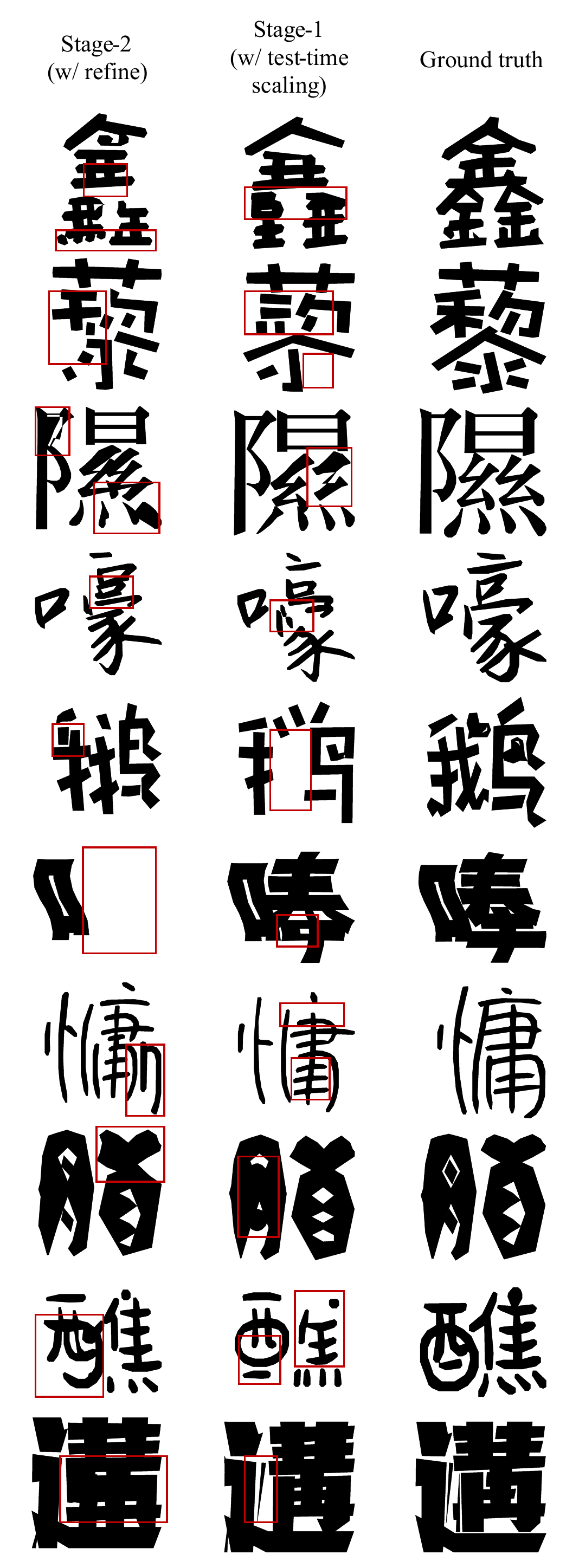}
    \caption{Examples of generation inaccuracies. The glyphs produced by our method may suffer from missing strokes, redundant strokes, and structural flaws, highlighted by red bounding boxes.}
  \label{fig:geni}
\end{figure}

\paragraph{Generation inaccuracies.} 
As shown in Fig.~\ref{fig:geni}, glyphs containing many short, adjacent strokes often lead to missing strokes, repeated transitions, or structural inaccuracies. 
This occurs because the model lacks a fine-grained understanding of how local strokes relate to the global glyph structure, making it difficult to model the dependencies between previously generated paths and the current one.

\end{document}